%% file: main.tex
\pdfoutput=1
\RequirePackage{amsmath}
\documentclass[runningheads]{llncs}
\usepackage[T1]{fontenc}
\usepackage{amssymb}
\usepackage{multirow}
\usepackage[table,xcdraw]{xcolor}
\usepackage{textcomp} 
\usepackage{subfig}
\usepackage{algorithmic}
\usepackage{algorithm}
\usepackage{graphicx}
\usepackage{hyperref}
\usepackage[normalem]{ulem}
\usepackage{multirow}
\usepackage{booktabs}
\usepackage{threeparttable}
\usepackage{siunitx}
\usepackage[misc]{ifsym}
\newcommand\doubleplus{+\kern-1.3ex+\kern0.8ex}
\begin{document}

\title{U-Net Inspired Transformer Architecture for \\Far Horizon Time Series Forecasting}

\titlerunning{Yformer for Far Horizon Time Series Forecasting}
%

\author{Kiran Madhusudhanan\inst{1}\Letter \and
Johannes Burchert \inst{1} \and
Nghia Duong-Trung \inst{2}\and
Stefan Born \inst{2} \and
Lars Schmidt-Thieme \inst{1} 
}
\authorrunning{K. Madhusudhanan et al.}
%
\institute{Institute for Computer Science, University of Hildesheim, Hildesheim, Germany \\
\email{\{madhusudhanan, burchert, schmidt-thieme\}@ismll.uni-hildesheim.de}
\and
Technische Universit\"at Berlin, Berlin, Germany \\
\email{nghia.duong-trung@tu-berlin.de, born@math.tu-berlin.de}}

\toctitle{U-Net Inspired Transformer Architecture for Far Horizon Time Series Forecasting}
\tocauthor{Kiran~Madhusudhanan, Johannes~Burchert, Nghia~Duong-Trung, Stefan~Born and Lars~Schmidt-Thieme}

\maketitle              
\begin{abstract}
Time series data is ubiquitous in research as well as in a wide variety of industrial applications. Effectively analyzing the available historical data and providing insights into the far future allows us to make effective decisions. Recent research has witnessed the superior performance of transformer-based architectures, especially in the regime of far horizon time series forecasting. However, the current state of the art sparse Transformer architectures fail to couple down- and upsampling procedures to produce outputs in a similar resolution as the input. We propose a U-Net inspired Transformer architecture named Yformer, based on a novel Y-shaped encoder-decoder architecture that (1) uses direct connection from the downscaled encoder layer to the corresponding upsampled decoder layer in a U-Net inspired architecture, (2) Combines the downscaling/upsampling with sparse attention to capture long-range effects, and (3) stabilizes the encoder-decoder stacks with the addition of an auxiliary reconstruction loss. Extensive experiments have been conducted with relevant baselines on three benchmark datasets, demonstrating an average improvement of 19.82, 18.41 percentage MSE and 13.62, 11.85 percentage MAE in comparison to the baselines for the univariate and the multivariate settings respectively.

\end{abstract}

\input{sections/introduction.tex}

\input{sections/relatedwork.tex}

\input{sections/background.tex}

\input{sections/methodology.tex}

\input{sections/experiments.tex}

\input{sections/ablation.tex}

\input{sections/conclusion.tex}



\bibliographystyle{splncs04}

\clearpage
\input{sections/appendix.tex}

\end{document}

%% file: sections/introduction.tex
\section{Introduction}

In the most simple case, time series forecasting deals with a scalar
time-varying signal and aims to predict or forecast its values in the near future; for example, countless applications in finance, healthcare, production automatization, etc. \cite{cao2018brits,sagheer2019time,sezer2020financial} can benefit from an accurate forecasting solution.
Often not just a single scalar signal is of interest, but multiple at once,
and further time-varying signals are available and even \textsl{known for the future}.
For example, suppose one aims to forecast the energy consumption of a house, it likely depends on the social time that one seeks to forecast for (such as the next hour or day), and also on features of these time points (such as weekday, daylight, etc.), which are known already for the future. This is also the case in model predictive control \cite{camacho2013model}, where one is interested
to forecast the expected value realized by some planned action, then this action is also known at the time of forecast.
More generally,
time series forecasting, nowadays deals with quadruples $(x,y,x',y')$
of known past predictors $x$, known past targets $y$, known future predictors $x'$
and sought future targets $y'$. 

\begin{figure}[ht]
\centering

\includegraphics[width=0.4\columnwidth]{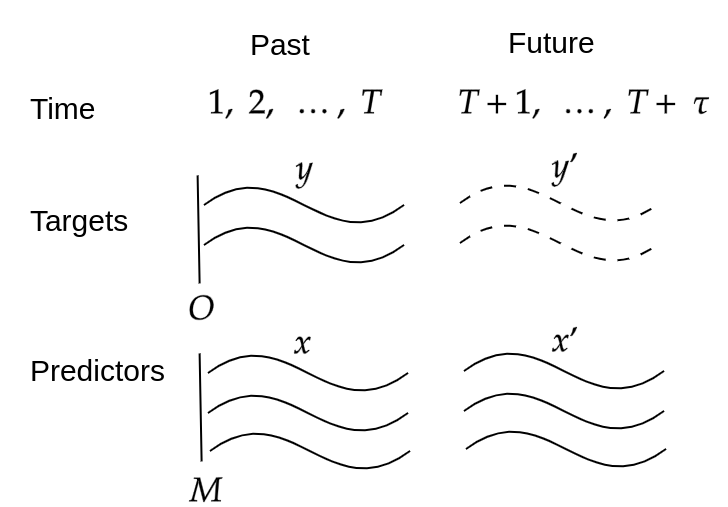}
\caption{General time series setting illustrating the quadruples $(x,y,x',y')$ denoting the \textsl{past predictors}, \textsl{past targets}, \textsl{future predictors} and \textsl{future targets} respectively. Given the history information $(x, y)$ until time $t = T$ and the future predictors $(x')$ for the next $\tau$ time steps, time series forecasting predicts the target $y'$ from $t = T+1, \dots, \tau$ time steps. In the figure, $O$ and $M$ represents the respective channels of the targets and the predictors.}
\label{fig:ts_ps}
\end{figure}

Time series problems can often be addressed by methods developed initially
for images, treating them as 1-dimensional images. Especially for
time-series classification many typical time series encoder architectures
have been adapted from models for images \cite{wang2017time,ZOU201939}. 
Time series forecasting then is closely related to image outpainting \cite{wang2019srn},
the task to predict how an image likely extends to the left, right, top or bottom,
as well as to the more well-known task of image segmentation,
where for each input pixel, an output pixel has to be predicted, whose channels
encode pixel-wise classes such as vehicle, road, pedestrian say for road scenes.
Time series forecasting combines aspects from both problem settings:
information about targets from shifted positions (e.g., the past targets $y$ as
 in image outpainting) and
information about other channels from the same positions (e.g., the future predictors $x'$
 as in image segmentation).
One of the most successful, principled architectures for the image segmentation
task are U-Nets introduced in \cite{ronneberger2015u}, an architecture that successively downsamples/coarsens
its inputs and then upsamples/refines the latent representation with
deconvolutions also using the latent representations of the same detail level,
tightly coupling down- and upsampling procedures and thus yielding latent
features on the same resolution as the inputs.

Following the great success in Natural Language Processing (NLP) applications, attention-based, esp. transformer-based
architectures \cite{vaswani2017attention} that model pairwise interactions
between sequence elements have been recently adapted for
time series forecasting. One of the significant
challenges, is that the length of the time series, are often one or two magnitudes of order larger than the (sentence-level) NLP problems. 

Plenty of approaches aim to mitigate the quadratic complexity $O(T^2)$ in
the sequence/time series length $T$ to at most $O(T\log T)$.
For example, the Informer architecture
\cite{zhou2020informer}, adapts the transformer with a sparse attention
mechanism and a successive downsampling/coarsening of the past time series. As in the original transformer, only the coarsest representation is fed
into the decoder. Possibly to remedy the loss in resolution by this procedure,
the Informer feeds its input a second time into the decoder network, this time
without any coarsening. 

While forecasting problems share many commonalities with image segmentation
problems, transformer-based architectures like the Informer do not
involve coupled down- and upscaling procedures to yield predictions
on the same resolution as the inputs. 
Thus, we propose a novel Y-shaped architecture that
\begin{enumerate}
\item Couples downscaling/upscaling to leverage both, coarse and fine-grained
       features for time series forecasting,
\item Combines the coupled scaling mechanism with sparse attention modules to capture long-range effects on all scale levels, and
\item Stabilizes encoder and decoder stacks by reconstructing the recent past.
\end{enumerate}

%% file: sections/relatedwork.tex
\section{Related Work}

\textbf{Time Series Forecasting}: While Convolutional Neural Network (CNN) and Recurrent Neural network (RNN) based architectures \cite{RangapuramDeepState,salinas2020deepar} outperform traditional methods like ARIMA \cite{BoxArima} and exponential smoothing methods \cite{hyndman2018forecasting}, the addition of attention layers \cite{vaswani2017attention} to model time series forecasting has proven to be very beneficial across different problem settings \cite{10.1145/3292500.3330662,lai2018modeling,qin2017dual,wu2021autoformer}. Attention allows direct pair-wise interaction with eccentric events (like holidays) and can model temporal dynamics inherently unlike RNNs and CNNs that fail to capture long-range dependencies directly. Recent work like Reformer \cite{kitaev2020reformer}, Linformer, \cite{wang2020linformer}, Triformer \cite{cirstea2022triformer} and Informer \cite{zhou2020informer} have focused on reducing the quadratic complexity of modeling pair-wise interactions to a lower complexity with the introduction of restricted attention layers. Consequently, they can predict for longer forecasting horizons but are hindered by their capability of aggregating features and maintaining the resolution required for far horizon forecasting. 

\textbf{U-Net}: The Yformer model is inspired by the famous U-Net architecture introduced in \cite{ronneberger2015u} originating from the field of medical image segmentation. The U-net architecture is capable of compressing information by aggregating over the inputs and up-sampling embeddings to the same resolutions as that of the inputs from their compressed latent features. While there exist U-Net based transformer architectures within the vision community \cite{petit2021u,zhou2021nnformer}, to the best of our knowledge U-Net based transformer architecture for time series forecasting remains unexplored. Current transformer architectures like the Informer \cite{zhou2020informer} do not utilize up-sampling techniques even though the network produces intermediate multi-resolution feature maps. Our work aims to capitalize on these multi-resolution feature maps and use the U-net shape effectively for the task of time series forecasting. In \cite{perslev2019u}, the authors have successfully applied U-Net architecture for the task of time series segmentation, illustrating superior results in the task. These motivate the use of a U-Net-inspired architecture for time series forecasting as current methods fail to couple sparse attention mechanism with the U-Net shaped architecture for time series forecasting.

\textbf{Reconstruction Loss}: Reconstruction loss is widely used in the domain of time series outlier detection \cite{tungbcc19} and is less popular within the Time Series Forecasting community. Although recent time series forecasting architecture like the N-Beats \cite{nbeats} tries to reconstruct part of the past time steps (backcasting) as an effective method to improve model performance, the majority of transformer-based time series forecasting architectures \cite{10.1145/3292500.3330662,lai2018modeling,tft} fail to utilize the reconstruction loss as an auxiliary target to improve performance. In \cite{jawed2019multi}, the authors demonstrate a multi-task approach for time series forecasting that couples an auxiliary task of predicting known channels along with the target channel for improved regularization. Additionally, recent studies \cite{le2018supervised} have shown that the addition of the reconstruction term to any loss function generally provides uniform stability and bounds on the generalization error, therefore leading to a more robust model overall with no negative effect on the performance.

%% file: sections/background.tex
\section{Problem Formulation}
By a \textbf{time series $x$ with $M$ channels}, we mean a finite sequence of
  vectors in $\mathbb{R}^M$,
denote their space by
  $\mathbb{R}^{*\times M}:= \bigcup_{T\in \mathbb{N}} \mathbb{R}^{T\times M}$,
and their length by $|x|:= T$ (for $x\in \mathbb{R}^{T\times M}, M\in\mathbb{N}$).
We write $(x,y)\in \mathbb{R}^{*\times (M+O)}$ to denote two time series
of same length with $M$ and $O$ channels for the predictors and targets, respectively. 
We model a \textbf{time series forecasting instance} as a quadruple $(x,y,x',y')\in\mathbb{R}^{*\times (M+O)}\times \mathbb{R}^{*\times (M+O)}$, where $x,y$ denote the past predictors and targets until a reference time point $T$ and $x',y'$ denote the future predictors and targets from the reference point $T$ to the next $\tau$ (forecast horizon) time steps.

For a \textbf{Time Series Forecasting Problem}, given (i) a sample $\mathcal{D}:=\{$ $(x_1,y_1,x'_1,y'_1),$ $\ldots,(x_N,y_N,x'_N,y'_N)\}$ from an unknown distribution $p$ of time series forecasting instances and (ii) a function $\ell: \mathbb{R}^{*\times (O+O)} \rightarrow\mathbb{R}$ called loss, we attempt to find a function $\hat y:\mathbb{R}^{*\times (M+O)}\times \mathbb{R}^{*\times M}\rightarrow \mathbb{R}^{*\times O}$ (with $|\hat y(x,y,x')|=|x'|$) with minimal expected loss

\begin{equation}
    \begin{aligned}
       \mathbb{E}_{(x,y,x',y')\sim p}\ \ell(y', \hat y(x,y,x'))
    \end{aligned}
\end{equation}

The loss $\ell$ usually is the mean absolute error (MAE) or mean squared error (MSE) averaged over future time points:

\begin{equation}
   \ell^{\text{mae}}(y', \hat y) :=
    \frac{1}{|y'|}\sum_{t=1}^{|y'|} \frac{1}{O} ||y'_t-\hat y_t||_1, \quad
    \ell^{\text{mse}}(y', \hat y) :=
    \frac{1}{|y'|}\sum_{t=1}^{|y'|} \frac{1}{O} ||y'_t-\hat y_t||_2^2
\label{eqn:msemae}
\end{equation}

Furthermore, if there is only one target channel and no predictor channels ($O=1, M=0$),
the time series forecasting problem is called \textbf{univariate}, otherwise \textbf{multivariate}.

\section{Background}

Our work incorporates restricted attention based Transformer in a U-Net inspired architecture. For this reason, we base our work on the current state of the art sparse attention model Informer, introduced in \cite{zhou2020informer}. We provide a brief overview of the \textsl{ProbSparse} attention and the \textsl{Contracting ProbSparse Self-Attention Blocks} used in the Informer model for completeness.

\textbf{\textsl{ProbSparse} Attention}: The \textsl{ProbSparse} attention mechanism restricts the canonical attention \cite{vaswani2017attention} by selecting a subset $u$ of dominant queries from available sequence length $L_Q$ having the largest variance  across all the keys. Consequently, the dense query matrix ${\boldsymbol{Q}} \in \mathbb{R}^{L_Q \times d}$ in the canonical attention is replaced by a sparse query matrix $\overline{\boldsymbol{Q}} \in \mathbb{R}^{L_Q \times d}$ consisting of the $u$ dominant queries. \textsl{ProbSparse} attention can hence be defined as: 

\begin{equation}
\begin{aligned}
  \mathcal{A^{\text{PropSparse}}}(\boldsymbol{\overline{Q}}, \boldsymbol{K}, \boldsymbol{V}) &= \text{Softmax}(\frac{\boldsymbol{\overline{Q}}\boldsymbol{K}^T}{\sqrt{d}})\boldsymbol{V}
\end{aligned}
\label{eqn:probsparseattn}
\end{equation}

where $d$ denotes the input dimension to the attention module. For more details on the \textsl{ProbSparse} attention mechanism, we refer the reader to \cite{zhou2020informer}.

\textbf{Contracting ProbSparse Self-Attention Blocks}: The Informer model uses \textsl{Contracting ProbSparse Self-Attention Blocks} to distill out redundant information from the long history input sequence $(x,y)$ in a pyramid structure motivated from the image domain \cite{lin2017feature}. The sequence of operations within a block begins with a \textsl{ProbSparse} self-attention that takes as input the hidden representation $h_i$ from the $i^{th}$ block and projects the hidden representation into query, key and value for self-attention. This is followed by convolution operations $(\operatorname{Conv1d})$ \cite{alexnet}, and finally the Max-Pooling ($\operatorname{MaxPool}$) \cite{alexnet} operation reduces the latent dimension by effectively distilling out redundant information at each block as summarized in Algorithm 1. Here, $\operatorname{ELU}$ represents the ELU activation function \cite{elu} and $\operatorname{LayerNorm}$ is the Layer Normalization operation \cite{layernorm}.
The encoder block in the Informer model \cite{zhou2020informer} stacks multiple \textsl{Contracting ProbSparse Self-Attention Block} blocks and produce multi-resolution encoder embeddings following a pyramid structure.

\begin{algorithm}[H]
\label{alg:contracting}
    \textit{Input}  : $h_i$  \\
    \textit{Output} : $h_{i+1}$\\
    $h_{i+1} \gets \operatorname{ProbSparseAttn} (h_i, h_i)$ \\
    $h_{i+1} \gets \operatorname{Conv1d} (h_{i+1})$\\
    $h_{i+1} \gets \operatorname{LayerNorm} (h_{i+1})$ \\
    $h_{i+1} \gets \operatorname{MaxPool}(\operatorname{ELU}(\operatorname{Conv1d} (h_{i+1})))$
 \caption{Contracting ProbSparse Self-Attention Block} 
\end{algorithm}

%% file: sections/methodology.tex
\section{Methodology}

\begin{figure}[!ht]
    \hspace*{-0.7cm} 
    \subfloat[Informer Architecture\label{fig:informer}]{%
      \includegraphics[width=0.55\textwidth]{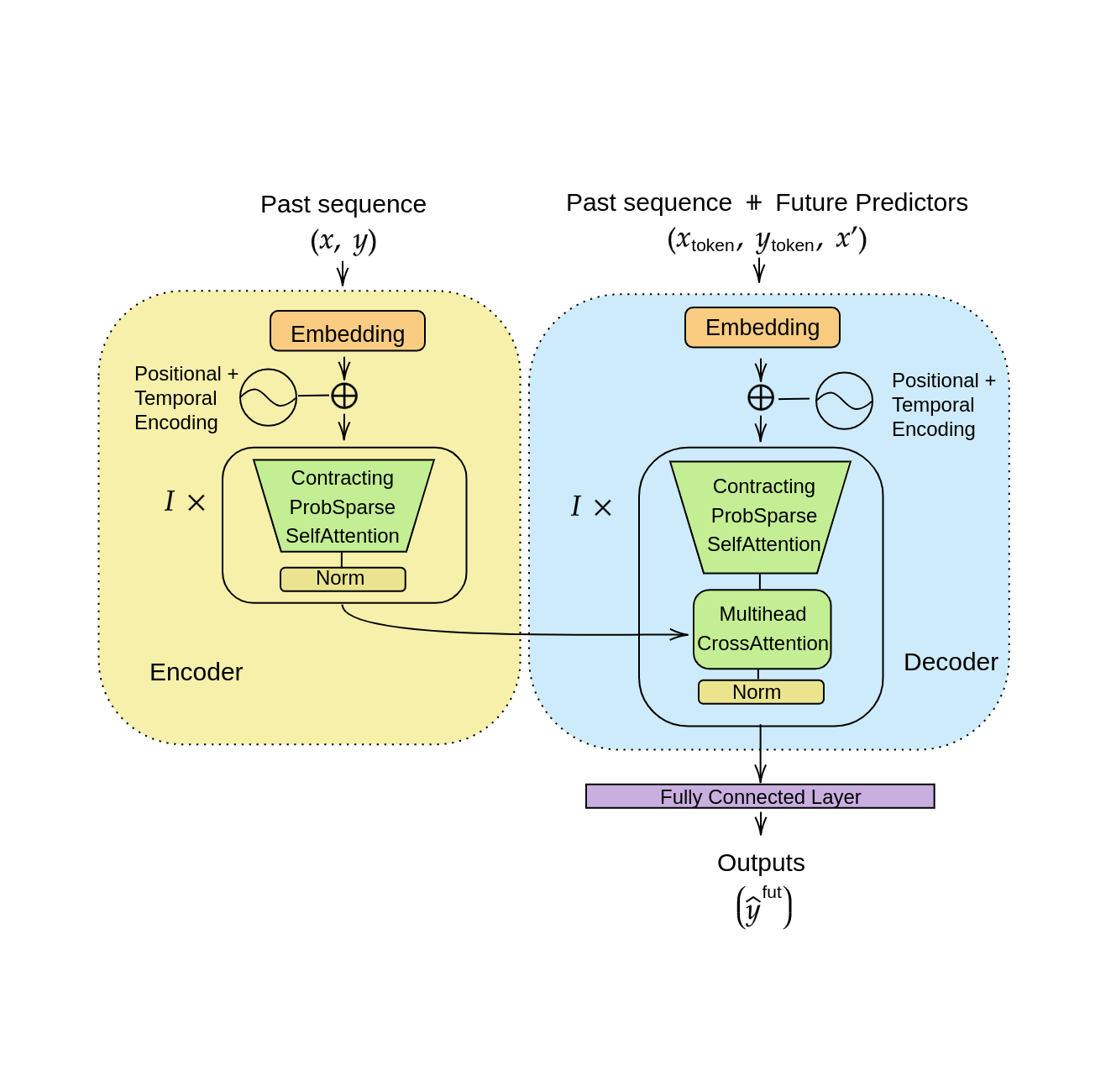}
    }
    \subfloat[Yformer Architecture\label{fig:Yformer}]{%
      \includegraphics[width=0.45\textwidth]{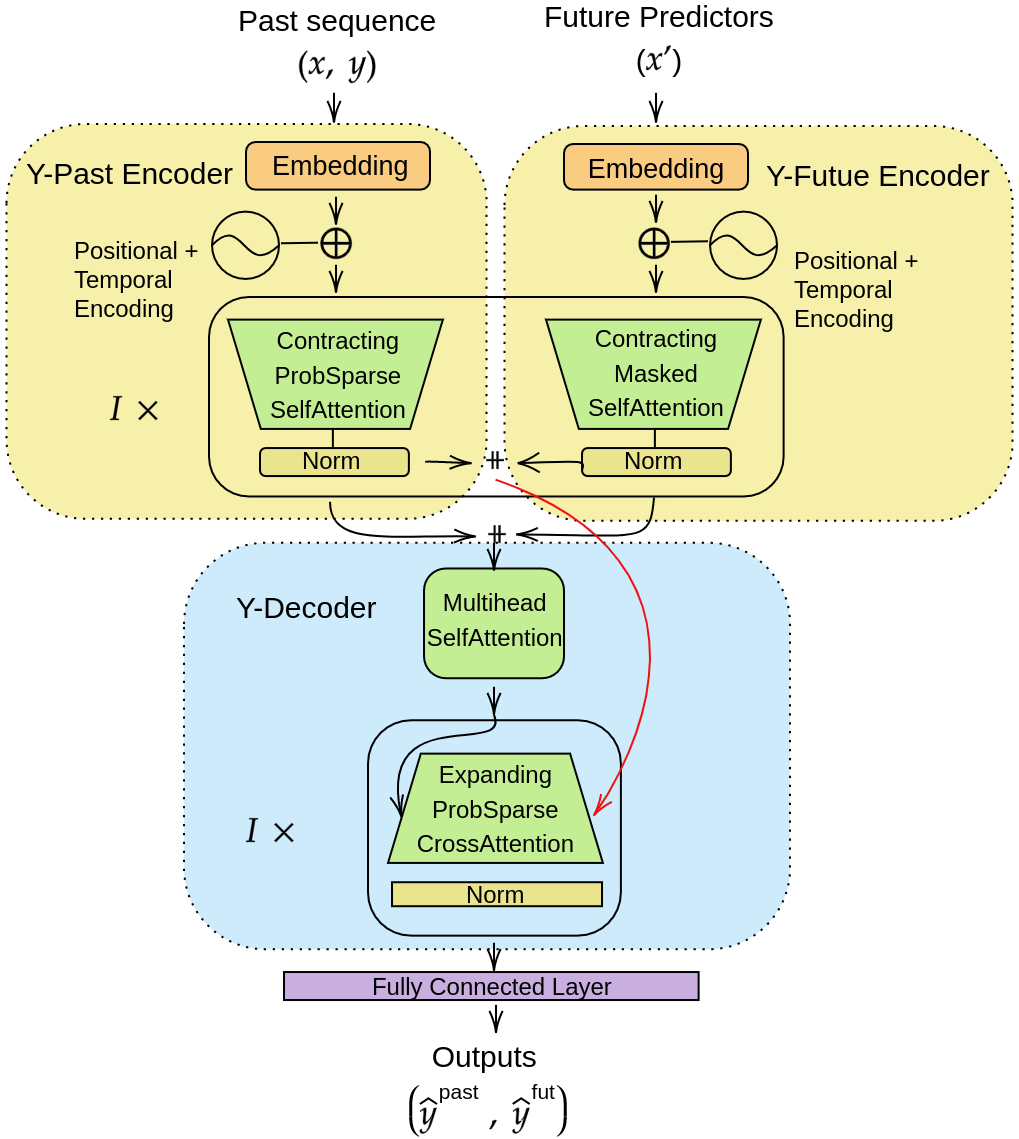}
    }
\caption{Comparison of Informer and Yformer architecture highlighting the three key differences. (1) The Informer architecture process part of the past input data $(x, y)$ within the decoder as $(x_{\text{token}}, y_{\text{token}})$ along with the future predictors $(x')$. The Yformer avoids this redundant reprocessing of $(x,y)$ and uses a masked self-attention network for embedding the only the future predictors $(x')$. (2) The Informer uses the final encoder embedding as the input to the decoder. The Yformer passes a concatenated ( $\doubleplus$) representation ($e_i$) of the $i^{\text{th}}$ Y-Past and Y-Future Encoder embedding to the ${I-i}^{th}$ layer of the Y-Decoder, forming a U-Net connection (represented in red) between the encoder and the decoder. (3) The Yformer architecture predicts both the input reconstruction $\hat{y}^{\text{past}}$ and future predictions $\hat{y}^{\text{fut}}$.}
\label{fig:architecture}
\end{figure}

\begin{figure}[t]
 \centering
 \includegraphics[width=0.6\textwidth]{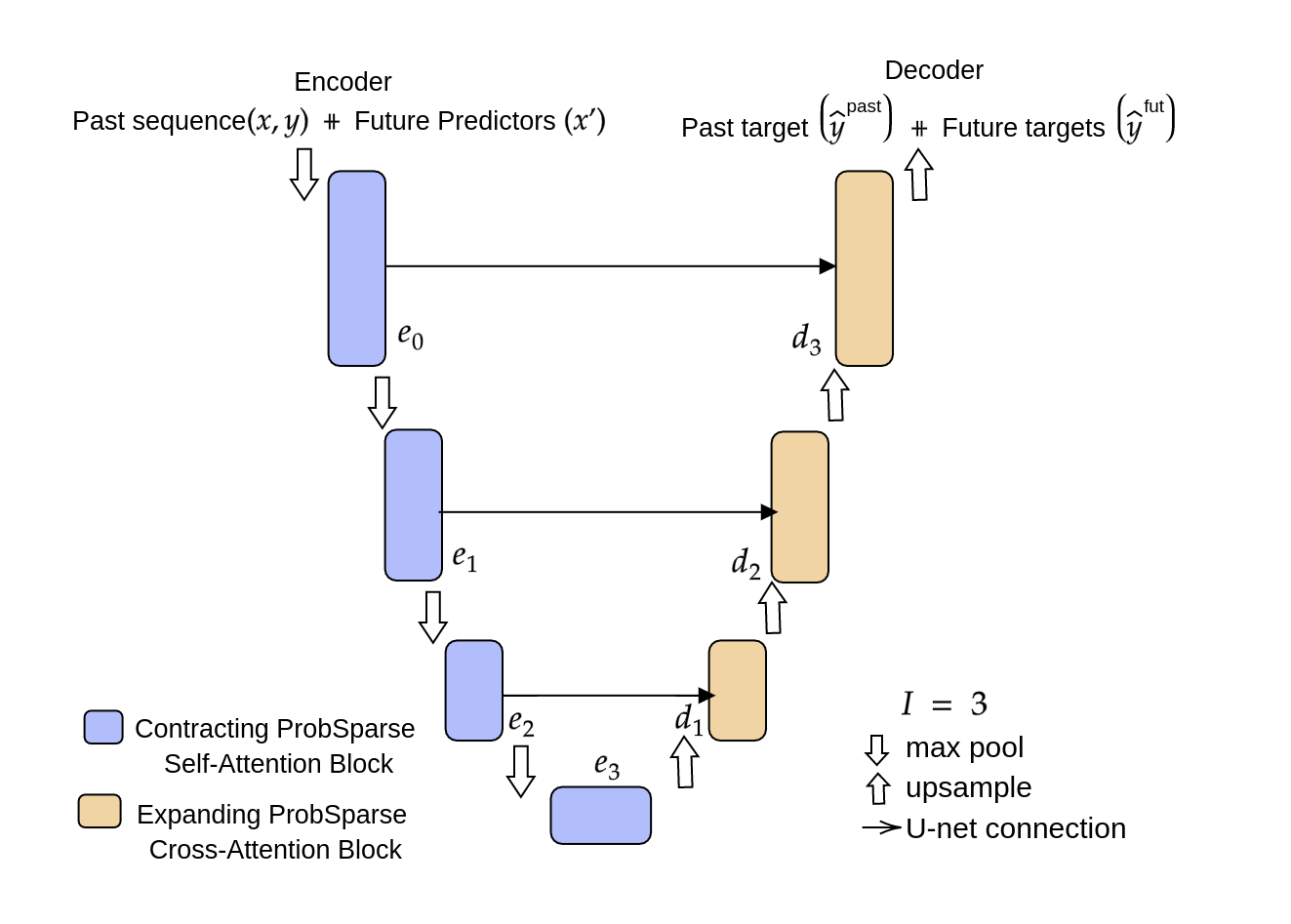}
 \caption{U-Net connections for effectively utilizing embeddings at multiple resolutions in the Yformer. The 
 Y-Past Encoder embeddings and the Y-Future Encoder embeddings are concatenated within the Yformer encoder. A direct connection is allowed between the contracting encoder embedding $(e_i)$ and the corresponding expanding decoder embedding $(d_{I-i})$. ($\doubleplus$ denotes concatenation) }
 \label{fig:unet}
\end{figure}

The Yformer model is a Y-shaped symmetric encoder-decoder architecture that is specifically designed to take advantage of the multi-resolution embeddings generated by the \textsl{Contracting ProbSparse Self-Attention Blocks}. The fundamental design consideration is the adoption of U-Net-inspired connections to extract encoder features at multiple resolutions and provide a direct connection to the corresponding symmetric decoder block. The Yformer additionally utilizes reconstruction loss to learn generalized embeddings that better approximate the data generating distribution. Figures \ref{fig:informer} and \ref{fig:Yformer} compares the Informer architecture with the Yformer and Figure \ref{fig:unet} illustrates the U-Net connections employed by the Yformer model.

The \textbf{Y-Past Encoder} of the Yformer is designed using a similar encoder structure as that of the Informer (Figure \ref{fig:informer}). The Y-Past Encoder embeds the past sequence $(x, y)$ into a scalar projection along with the addition of positional and temporal embeddings. Multiple \textsl{Contracting ProbSparse Self-Attention Blocks} are used to generate encoder embeddings at various resolutions following a contracting pyramid structure. The Informer model uses the final low-dimensional embedding as the input to the decoder whereas, the Yformer retains the embeddings at multiple resolutions to be passed on to the decoder. This allows the Yformer to use high-dimensional lower-level embeddings effectively.

The \textbf{Y-Future Encoder} of the Yformer mitigates the redundant reprocessing of the past sequence $(x,y)$ (used as tokens $(x_{\text{token}}, y_{\text{token}})$ in the Informer architecture) by passing only the future predictors $(x')$ through the Y-Future Encoder and utilizing the multi-resolution embeddings to dismiss the need for tokens entirely. The attention blocks in the Y-Future encoder are based on a masked canonical self-attention mechanism \cite{vaswani2017attention} to prevent any information leak from the future time steps into the past. Thus, the Y-Future Encoder is designed by stacking multiple \textsl{Contracting ProbSparse Self-Attention Blocks} where the \textsl{ProbSparse} attention is replaced by the \textsl{Masked Attention}. We name these blocks \textsl{Contracting Masked Self-Attention Blocks}.

The Yformer processes the past inputs and the future predictors separately within its encoders. However, considering the time steps, the future predictors are a continuation of the past time steps. For this reason, the Yformer model concatenates ( represented by the symbol $\doubleplus$) the past encoder embedding and the future encoder embedding along the time dimension after each encoder block, preserving the continuity between the past input time steps and the future time steps. Let $i$ represent the index of an encoder block, then $e^{\text{past}}_{i+1}$ and $e^{\text{fut}}_{i+1}$ represent the output from the past encoder and the future encoder respectively. The final concatenated encoder embedding $(e_{i+1})$ is calculated as, 

\begin{equation}
\begin{aligned}
    e^{\text{past}}_{i+1} &= \operatorname{ContractingProbSparseSelfAttentionBlock}(e^{\text{past}}_{i}) \\
    e^{\text{fut}}_{i+1} &= \operatorname{ContractingMaskedSelfAttentionBlock}(e^{\text{fut}}_{i}) \\
    e_{i+1} &= e^{\text{past}}_{i+1} \doubleplus e^{\text{fut}}_{i+1}
    \label{Eq:concatencoders}
\end{aligned}
\end{equation}
The encoder embeddings represented by $\mathcal{E} = [e_0, \dots, e_{I}]$ (where $I$ is the number of encoder layers) contain the combination of past and future embeddings at multiple resolutions.

The \textbf{Y-Decoder} of the Yformer consists of two parts. The first part takes as input the final concatenated low-dimensional embedding $(e_I)$ of the encoders and performs a multi-head canonical self-attention mechanism. Since the canonical self-attention layer is separated from the repeating attention blocks within the decoder, the Yformer complexity from this full attention module does not increase with an increase in the number of decoder blocks. The U-Net architecture inspires the second part of the Y-Decoder. Consequently, the decoder is structured in a symmetric expanding path identical to the contracting encoder (Figure \ref{fig:unet}). We realize this idea by introducing  \textsl{Expanding ProbSparse Cross-Attention Block} for symmetric upsampling.

The \textbf{\textsl{Expanding ProbSparse Cross-Attention Block}} within the Yformer decoder performs two tasks: (1) upsample the compressed encoder embedding $e_I$ and (2) perform restricted cross attention between the expanding decoder embedding $d_{I-i}$ and the corresponding encoder embedding $e_i$ as shown below. 
\begin{algorithm}[H]
    \textit{Input}  : $d_{I-i}, e_i$  \\
    \textit{Output} : $d_{I-i+1}$\\
    $d_{I-i+1} \gets \operatorname{ProbSparseCrossAttn} (d_{I-i}, e_i)$ \\
    $d_{I-i+1} \gets \operatorname{Conv1d} (d_{I-i+1})$\\
    $d_{I-i+1} \gets \operatorname{LayerNorm} (d_{I-i+1})$ \\
    $d_{I-i+1} \gets \operatorname{ELU}(\operatorname{ConvTranspose1d} (d_{I-i+1})))$
 \caption{Expanding ProbSparse Cross-Attention Block}
\end{algorithm}

The \textsl{Expanding ProbSparse Cross-Attention Blocks} within the Yformer decoder uses a $\operatorname{ProbSparseCrossAttn}$ to construct direct connections between the lower levels of the encoder and the corresponding symmetric higher levels of the decoder. Direct connections from the encoder to the decoder are an essential component for the majority of models within the image domain. For example, ResNet \cite{he2016deep}, and DenseNet \cite{huang2017densely} have demonstrated that direct connections between previous feature maps, strengthen feature propagation, reduce parameters, mitigate vanishing gradients and encourage feature reuse. However, current transformer-based architectures fail to utilize these direct connections.  

We utilize $\operatorname{ConvTranspose1d}$ or popularly known as $\operatorname{Deconvolution}$ for incrementally increasing the embedding space. The famous U-Net architecture uses a symmetric expanding path using such $\operatorname{Deconvolution}$ layers. This property enables the model to not only aggregate over the input but also upscale the latent dimensions, improving the overall expressivity of the architecture. The decoder of Yformer follows a similar strategy by employing $\operatorname{Deconvolution}$ to expand the embedding space of the encoded output as shown in Figure \ref{fig:unet}.

Finally, a fully connected layer ($\operatorname{LinearLayer}$) predicts the future time steps $\hat{y}^{\text{fut}}$ from the  final decoder layer $(d_I)$ and additionally reconstructs the past input targets $\hat{y}^{\text{past}}$ for the reconstruction auxiliary loss.

\begin{equation}
    \begin{aligned}
        {[\hat{y}^{\text{past}}, \hat{y}^{\text{fut}}]} &= \operatorname{LinearLayer} (d_I) \\
    \end{aligned}
    \label{eqn:prediction}
\end{equation}

The addition of reconstruction loss to the Yformer as an auxiliary loss serves two significant purposes. Firstly, the reconstruction loss acts as a data-dependent regularization term that reduces overfitting by learning embeddings that are more general \cite{Jarrett2020Target-Embedding}. Secondly, the reconstruction loss helps in producing future output in a similar distribution as the inputs. For far horizon forecasting, we are interested in learning a future-output distribution, however, the future-output distribution and the past-input distribution arise from the same data generating process. Therefore having an auxiliary reconstruction loss would direct the gradients to a better approximate of the data generating process. Consequently, the Yformer model is trained on the combined loss $\ell$, 

\begin{equation}
    \ell = \alpha \, \ell^{\text{mse}}(y, \hat{y}^{\text{past}}) + (1-\alpha) \, \ell^{\text{mse}}(y', \hat{y}^{\text{fut}}) 
    \label{eqn:reconstruction}
\end{equation}
where the first term tries to learn the past targets $y$ and the second term learns the future targets $y'$. We use the reconstruction factor $(\alpha)$ to vary the importance of reconstruction and future prediction and tune this as a hyperparameter.

%% file: sections/experiments.tex
\section{Experiments}

\subsection{Datasets}
We compare the experimental results of our proposed YFormer architecture, with that of the Informer on three real-world public datasets.

\textbf{ETTh1 and ETTh2} (Electricity Transformer Temperature\footnote{https://
github.com/zhouhaoyi/ETDataset.}):
These real-world datasets for the electric power deployment introduced by \cite{zhou2020informer} combine short-term periodical patterns, long-term periodical patterns, long-term trends, and irregular patterns. The data consists of load and temperature readings from two transformers at two different stations with varying load conditions. The ETTm1 dataset is generated by splitting ETTh1 dataset into 15-minute intervals. The dataset has six features and 70,080 data points in total. For easy comparison, we kept the splits for train/val/test consistent with the published results in \cite{zhou2020informer}, where the available 20 months of data is split as 12/4/4. For the Univariate setting, 'OT' (Oil Temperature) was set as the target value.

\textbf{ECL} (Electricity Consuming Load\footnote{https://archive.ics.uci.edu/ml/
datasets/ElectricityLoadDiagrams20112014}): This electricity dataset represents the electricity consumption from 2011 to 2014 of 370 clients recorded in 15-minutes periods in Kilowatt (kW). We split the data into 15/3/4 months for train, validation, and test respectively as in \cite{zhou2020informer}. For the Univariate setting, 'MT 320' was set as the target value.

\subsection{Experimental Setup}
\textbf{Baseline}: Our main baseline is the Informer architecture. As a second baseline, we also compare the second-best performing model which is the Informer that uses canonical attention module \cite{zhou2020informer} represented as Informer$^{\dag}$. Furthermore, we also compare against DeepAR \cite{salinas2020deepar}, and LogTrans \cite{li2019enhancing} for the univariate setting, and LSTnet \cite{lai2018modeling} for the multivariate setting as they outperform the Informer baseline for certain forecasting horizons. For a quick analysis, we present the percent improvement achieved by the Yformer over the current best results as the final column in Tables \ref{tbl:univariate}, \ref{tbl:multivariate}.

For a fair comparison, we retain the design choices from the Informer baseline like the history input length $(T)$ for a particular forecast length ($\tau$), so that any performance improvement can exclusively be attributed to the architecture of the Yformer model and not to an increased history input length. We performed a grid search for learning rates of $\{0.001, 0.0001\}$, $\alpha$-values of $\{0, 0.3, 0.5, 0.7, 1\}$, number of encoder and decoder blocks  $I = \{2, 3, 4\}$ while keeping all the other hyperparameters the same as the Informer. Furthermore, Adam optimizer and an early stopping criterion with a patience of three epochs was used for all experiments. To counteract overfitting, we tried dropout with varying ratios but interestingly found the effect to be minimal in the results. Therefore, we adopt weight-decay for our experiments with factors $\{0, 0.02, 0.05\}$ for additional regularization. We select the optimal hyperparameters based on the lowest validation loss. 

\begin{table*}[t]
    \caption{Univariate results for three datasets (four cases) with different prediction lengths $\tau \in \{24,48,96,168,288,336,672,720,960\}$.}
    \centering
    \fontsize{9pt}{9pt}\selectfont
    \centering
    \resizebox{1\textwidth}{!}{%
    \begin{tabular}{c|c|c|c|c|c|c|SS|}
    \toprule[1.0pt]
    \multicolumn{2}{c|}{Methods}    & {Yformer}     & {Informer} & {Informer$^{\dag}$} & {LogTrans}  & {DeepAR} & \multicolumn{2}{c|}{Improvement\%} \\
    \midrule[0.5pt]
    \multicolumn{2}{c|}{Metric}     & MSE~~MAE     & MSE~~MAE               & MSE~~MAE                    & MSE~~MAE           & MSE~~MAE            & {MSE}            & {MAE}\\
    \midrule[1.0pt]
    \multirow{5}{*}{\rotatebox{90}{ETTh$_1$}}      
         &     24   &    \textbf{0.082}~~\textbf{0.230} &   0.098~~0.247    &   \underline{0.092}~~\underline{0.246}     &   0.103~~0.259    &   0.107~~0.280    &   10.87 & 6.50 \\
         &     48   &    \textbf{0.139}~~\textbf{0.308} &   \underline{0.158}~~\underline{0.319}     &   0.161~~0.322    &   0.167~~0.328    &   0.162~~0.327    &   12.03 & 3.45 \\
         &     168  &    \textbf{0.111}~~\textbf{0.268} &   \underline{0.183}~~\underline{0.346}     &   0.187~~0.355    &   0.207~~0.375    &   0.239~~0.422    &   39.34 & 22.54 \\
         &     336  &    \textbf{0.195}~~\textbf{0.365} &   0.222~~0.387    &   \underline{0.215}~~\underline{0.369}     &   0.230~~0.398    &   0.445~~0.552    &   09.30 & 1.08 \\
         &     720  &    \textbf{0.226}~~\textbf{0.394}&    0.269~~0.435    &   \underline{0.257}~~\underline{0.421}     &   0.273~~0.463    &   0.658~~0.707    &   12.06 & 6.41 \\
    \midrule[0.5pt]
    \multirow{5}{*}{\rotatebox{90}{ETTh$_2$}}      
         &     24   &    \textbf{0.082}~~\textbf{0.221} &   \underline{0.093}~~\underline{0.240}     &   0.099~~0.241    &   0.102~~0.255     &  0.098~~0.263    &   11.83   & 7.92 \\
         &     48    &   0.172~~0.334 &      \textbf{0.155}~~\textbf{0.314}      &   \underline{0.159}~~\underline{0.317}     &   0.169~~0.348     &  0.163~~0.341    &   -10.97  & -6.37 \\
         &     168   &   \textbf{0.174}~~\textbf{0.337} &   \underline{0.232}~~\underline{0.389}     &   0.235~~0.390    &   0.246~~0.422     &  0.255~~0.414    &   25.00   & 13.37 \\
         &     336   &   \textbf{0.224}~~\textbf{0.391} &   0.263~~\underline{0.417}  &   \underline{0.258}~~0.423  &   0.267~~0.437     &  0.604~~0.607    &   13.18   & 6.24 \\
         &     720   &   \textbf{0.211}~~\textbf{0.382} &   \underline{0.277}~~\underline{0.431}     &   0.285~~0.442    &   0.303~~0.493     &  0.429~~0.580    &   23.83   & 11.37 \\
    \midrule[0.5pt]
    \multirow{5}{*}{\rotatebox{90}{ETTm$_1$}}      
         &     24    &   \textbf{0.024}~~\textbf{0.118} &   \underline{0.030}~~\underline{0.137}     &   0.034~~0.160    &   0.065~~0.202    &   0.091~~0.243    &   20.00   & 13.87 \\
         &     48    &   \textbf{0.048}~~\textbf{0.173} &   0.069~~0.203    &   \underline{0.066}~~\underline{0.194}     &   0.078~~0.220    &   0.219~~0.362    &   27.27   & 10.82 \\
         &     96    &   \textbf{0.143}~~\textbf{0.311} &   0.194~~\underline{0.372}  &   \underline{0.187}~~0.384  &   0.199~~0.386    &   0.364~~0.496    &   23.53   & 16.40 \\
         &     288   &   \textbf{0.150}~~\textbf{0.316}&    \underline{0.401}~~0.554  &   0.409~~\underline{0.548}  &   0.411~~0.572    &   0.948~~0.795    &   62.59   & 42.34 \\
         &     672   &   \textbf{0.305}~~\textbf{0.476} &   \underline{0.512}~~\underline{0.644}     &   0.519~~0.665    &   0.598~~0.702    &   2.437~~1.352    &   40.43   & 26.09 \\
    \midrule[0.5pt]
    \multirow{5}{*}{\rotatebox{90}{ECL}}      
         &     48    &   \textbf{0.194}~~\textbf{0.322} &   0.239~~0.359    &   0.238~~0.368    &   0.280~~0.429    &   \underline{0.204}~~\underline{0.357}     &   4.90    & 9.80 \\
         &     168   &   \textbf{0.260}~~\textbf{0.361} &   0.447~~0.503    &   0.442~~0.514    &   0.454~~0.529    &   \underline{0.315}~~\underline{0.436}     &   17.46   & 17.20 \\
         &     336   &   \textbf{0.269}~~\textbf{0.375} &   0.489~~0.528    &   0.501~~0.552    &   0.514~~0.563    &   \underline{0.414}~~\underline{0.519}     &   35.02   & 27.75 \\
         &     720   &   \textbf{0.427}~~\textbf{0.479} &   \underline{0.540}~~\underline{0.571}     &   0.543~~0.578    &   4.891~~4.047    &   0.563~~0.595    &   20.93   & 19.50 \\
         &     960   &   0.595~~\textbf{0.573} &  \textbf{0.582}~~\underline{0.608}   &   0.594~~0.638    &   7.019~~5.105    &   0.657~~0.683    &   -2.23   & 16.11 \\
    \midrule[0.5pt]
    \multicolumn{2}{c|}{Count}      & {37}     & {3}                 & {0}                          & {0}                & {0}           & {}  & {}    \\
    \midrule[0.5pt]
    \multicolumn{2}{c|}{Average}      & \multicolumn{1}{c}{}     & \multicolumn{1}{c}{}                 & \multicolumn{1}{c}{}           & \multicolumn{1}{c}{}     & \multicolumn{1}{c|}{}           & {19.82}  & {13.62}    \\
    \bottomrule[1.0pt]
    
    \end{tabular}
    }
    \label{tbl:univariate}
    \end{table*}

For easy comparison, we choose two commonly used metrics for time series forecasting to evaluate the Yformer architecture, the MAE and MSE in Equation \ref{eqn:msemae}. We performed our experiments on GeForce RTX 2080 Ti GPU nodes with 32 GB ram and provide results as an average of three runs. The source code \footnote{https://github.com/18kiran12/Yformer-Time-Series-Forecasting} and optimal hyperparameter configurations are made public for reproducibility.

\subsection{Results and Analysis}

This section compares our results with the results reported in the Informer baseline both in uni- and multivariate settings for the multiple datasets and horizons. A direct comparison with the reported results \cite{zhou2020informer} is possible as the experimental setup and the problem settings are kept the same. The best-performing and the second-best models are highlighted in bold and in underline, respectively. 

\begin{table*}[t]
\caption{Multivariate results for three datasets (four cases) with different prediction lengths $\tau \in \{24,48,96,168,288,336,672,720,960\}$.}
    \centering
    \resizebox{1\textwidth}{!}{%
    \fontsize{9pt}{9pt}\selectfont
    \begin{tabular}{c|c|cc|cc|cc|cc|cc|SS|}
    \toprule[1.0pt]
    \multicolumn{2}{c}{Methods}   & \multicolumn{2}{|c}{Yformer}  & \multicolumn{2}{|c}{Informer} & \multicolumn{2}{|c}{Informer$^{\dag}$} & \multicolumn{2}{|c}{LogTrans}  & \multicolumn{2}{|c}{LSTnet}  & \multicolumn{2}{|c|}{Improvement\%}     \\
    \midrule[0.5pt]
    \multicolumn{2}{c|}{Metric}     & MSE                & MAE    & MSE                & MAE               & MSE                          & MAE              & MSE               & MAE             & MSE            & MAE      & {MSE}            & {MAE}        \\
    \midrule[1.0pt]
    \multirow{5}{*}{\rotatebox{90}{ETTh$_1$}} 
    & 24  &\textbf{0.485} & \textbf{0.492} & \underline{0.577}          & \underline{0.549}          & 0.620                   & 0.577                   & 0.686                   & 0.604                   & 1.293                   & 0.901         &15.94 & 10.38         \\
    & 48  &\textbf{0.530} & \textbf{0.537}  & \underline{0.685}          & \underline{0.625}          & 0.692                   & 0.671                   & 0.766                   & 0.757                   & 1.456                   & 0.960        &22.63 & 14.08         \\
    & 168 &\textbf{0.866} & \textbf{0.684} & \underline{0.931}          & \underline{0.752}          & 0.947                   & 0.797                   & 1.002                   & 0.846                   & 1.997                   & 1.214         &06.98 & 09.04         \\
    & 336 &\textbf{1.041} & \textbf{0.803} & 1.128                   & 0.873                   & \underline{1.094}          & \underline{0.813}          & 1.362                   & 0.952                   & 2.655                   & 1.369         &04.84 & 01.23         \\
    & 720 &\textbf{1.098} & \textbf{0.803} & \underline{1.215}          & \underline{0.896}          & 1.241                   & 0.917                   & 1.397                   & 1.291                   & 2.143                   & 1.380         &09.63 & 10.38        \\
    \midrule[0.5pt]
    \multirow{5}{*}{\rotatebox{90}{ETTh$_2$}} 
    & 24  &\textbf{0.412} & \textbf{0.498} & \underline{0.720}          & \underline{0.665}          & 0.753                   & 0.727                   & 0.828                   & 0.750                   & 2.742                   & 1.457        &42.78 &  25.11         \\
    & 48  &\textbf{1.171} & \textbf{0.865} & \underline{1.457}          & \underline{1.001}          & {1.461}                   & 1.077                   & 1.806                   & 1.034                   & 3.567                   & 1.687        &19.63 &  13.59        \\
    & 168 &\textbf{2.171} & \textbf{1.218} & 3.489                   & \underline{1.515}          & 3.485                   & 1.612                   & 4.070                   & 1.681                   & \underline{3.242}          & 2.513        &33.04 &  19.60        \\
    & 336 &\textbf{2.260} & \textbf{1.283} & 2.723                   & 1.340                   & 2.626                   & \underline{1.285}          & 3.875                   & 1.763                   & \underline{2.544}          & 2.591        &11.16 &  0.16        \\
    & 720 &\textbf{2.595} & \textbf{1.337} & \underline{3.467}          & \underline{1.473}          & 3.548                   & 1.495                   & 3.913                   & 1.552                   & 4.625                   & 3.709        &25.15 &  9.23       \\
    \midrule[0.5pt]
    \multirow{5}{*}{\rotatebox{90}{ETTm$_1$}} 
    & 24 &\textbf{0.289} &\textbf{0.363}  & 0.323                   & \underline{0.369}          & \underline{0.306}          & 0.371                   & 0.419                   & 0.412                   & 1.968                   & 1.170         &05.56  &  1.63        \\
    & 48 &\underline{0.486} &\textbf{0.457}  & 0.494                   & 0.503                   & \textbf{0.465}          & \underline{0.470}          & 0.507                   & 0.583                   & 1.999                   & 1.215         &-4.52 &   2.77       \\
    & 96  &\textbf{0.569} &\textbf{0.567} & \underline{0.678}          & 0.614                   & 0.681                   & \underline{0.612}          & 0.768                   & 0.792                   & 2.762                   & 1.542         &16.08 &   7.35       \\
    & 288 &\textbf{0.649} &\textbf{0.593} & \underline{1.056}                   & \underline{0.786}          & 1.162                   & 0.879                   & 1.462                   & 1.320                   & 1.257          & 2.076         &38.54 &   24.55       \\
    & 672 &\textbf{0.772} &\textbf{0.656} & \underline{1.192}          & \underline{0.926}          & 1.231                   & 1.103                   & 1.669                   & 1.461                   & 1.917                   & 2.941         &35.23 &   29.16       \\
    \midrule[0.5pt]
    \multirow{5}{*}{\rotatebox{90}{ECL}}      
    & 48  &\textbf{0.306} &\textbf{0.390} & 0.344                   & \underline{0.393}          & \underline{0.334}          & 0.399                   & 0.355                   & 0.418                   & 0.369                   & 0.445         &08.38 &   0.76       \\
    & 168 &\textbf{0.317} &\textbf{0.387} & 0.368                   & 0.424          & \underline{0.353}          & \underline{0.420}                   & 0.368                   & 0.432                   & 0.394                   & 0.476         &10.20 &   7.86       \\
    & 336 &\textbf{0.323} &\textbf{0.394} & 0.381                   & \underline{0.431}          & 0.381                   & 0.439                   & \underline{0.373}          & 0.439                   & 0.419                   & 0.477         &15.22 &   8.58       \\
    & 720 &\textbf{0.312} &\textbf{0.384} & 0.406                   & 0.443                   & \underline{0.391}          & \underline{0.438}          & 0.409                   & 0.454                   & 0.556                   & 0.565         &20.20 &   12.33       \\
    & 960 &\textbf{0.315} &\textbf{0.388} & \underline{0.460}          & \underline{0.548}          & 0.492                   & 0.550                   & 0.477                   & 0.589                   & 0.605                   & 0.599         &31.52 &   29.20      \\
    \midrule[1.0pt]
    \multicolumn{2}{c}{Count}       & \multicolumn{2}{|c}{39}  & \multicolumn{2}{|c}{0}                 & \multicolumn{2}{|c}{1}                          & \multicolumn{2}{|c}{0}           & \multicolumn{2}{|c}{0}  & \multicolumn{2}{|c|}{}   \\
    \midrule[0.5pt]
    \multicolumn{2}{c|}{Average}      & \multicolumn{1}{c}{}     & \multicolumn{1}{c}{}                 & \multicolumn{1}{c}{}                          & \multicolumn{1}{c}{}            & \multicolumn{1}{c}{}     & \multicolumn{1}{c}{}    & \multicolumn{1}{c}{}   & \multicolumn{1}{c}{}    & \multicolumn{1}{c}{}    & \multicolumn{1}{c|}{}   & {18.41}  & {11.85}    \\
    \bottomrule[1.0pt]
    \end{tabular}
    }
    
    \label{tbl:multivariate}
\end{table*}

\textbf{Univariate}: The proposed Yformer model is able to outperform the Informer baseline in 37 out of the 40 available tasks across different datasets and horizons by an average of 19.82\% MSE and 13.62 \% of MAE. Table \ref{tbl:univariate} illustrates that the superiority of the Yformer is not just limited to a far horizon but even for the shorter horizons and in general across datasets.  Considering the individual datasets, the Yformer surpasses the baselines by 8, 6.8, 21.9, and 18.1\% of MAE for the ETTh1, ETTh2, ETTm1, and ECL datasets respectively. MSE results illustrates an improvement of 16.7, 12.6, 34.8, and 15.2\% for the ETTh1, ETTh2, ETTm1, and ECL datasets respectively. We observe that the MAE for the model is greater at horizon 48 than the MAE at horizon 168 for the ETTh1 dataset. This may be a case where the reused hyperparameters from the Informer paper are far from optimal for the Yformer. The other results show consistent behavior of increasing error with increasing horizon length $\tau$. Additionally, this behavior is also observed in the Informer baseline for ETTh2 dataset (Table \ref{tbl:multivariate}), where the loss is 1.340 for horizon 336 and 1.515 for a horizon of 168. 

\textbf{Multivariate}: We observe a similar trend in the multivariate setting. Here the Yformer model outperforms the baseline method in almost all of the 40 tasks across the three datasets by a margin of 18.41 \% MSE and 11.85\% of MAE. There is a clear superiority of the proposed approach, especially for the longer horizons. Across the different datasets, the Yformer improves on the baseline results by 9, 13.5, 13.1, and 11.7\% of MAE, and 12, 26.3, 13.9, and 17.1\% of MSE  for the ETTh1, ETTh2, ETTm1, and ECL datasets respectively. We attribute the improvement in performance to superior architecture and the ability to approximate the data distribution due to the addition of auxiliary loss.

%% file: sections/ablation.tex
\section{Ablation study}
\label{sec:ablation}

Additional experiments were performed on the ETTm1 datasets to analyze the different components of the Yformer model. Similar ablation experiment results for ETTh2 dataset are reported in the Appendix section for reference.

\subsection{Y-former architecture}

\begin{figure}[t]
    \centering
    \begin{tabular}{c}
    
    \subfloat[ETTm1 Univariate\label{fig:ablation_archi_uni}]{%
      \includegraphics[width=0.40\textwidth]{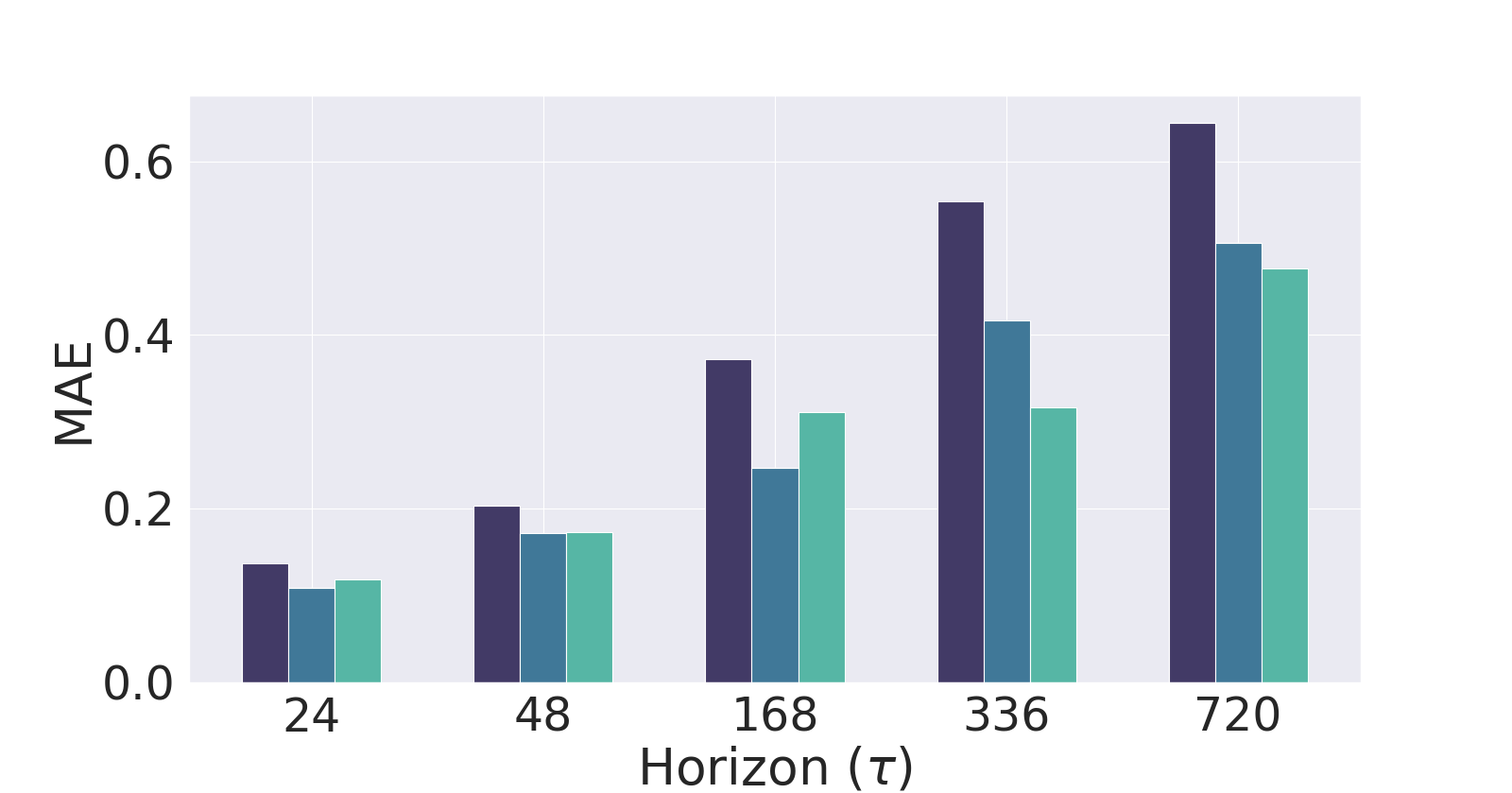}
    }

    \subfloat[ETTm1 Multivariate\label{fig:ablation_archi_multi}]{%
      \includegraphics[width=0.40\textwidth]{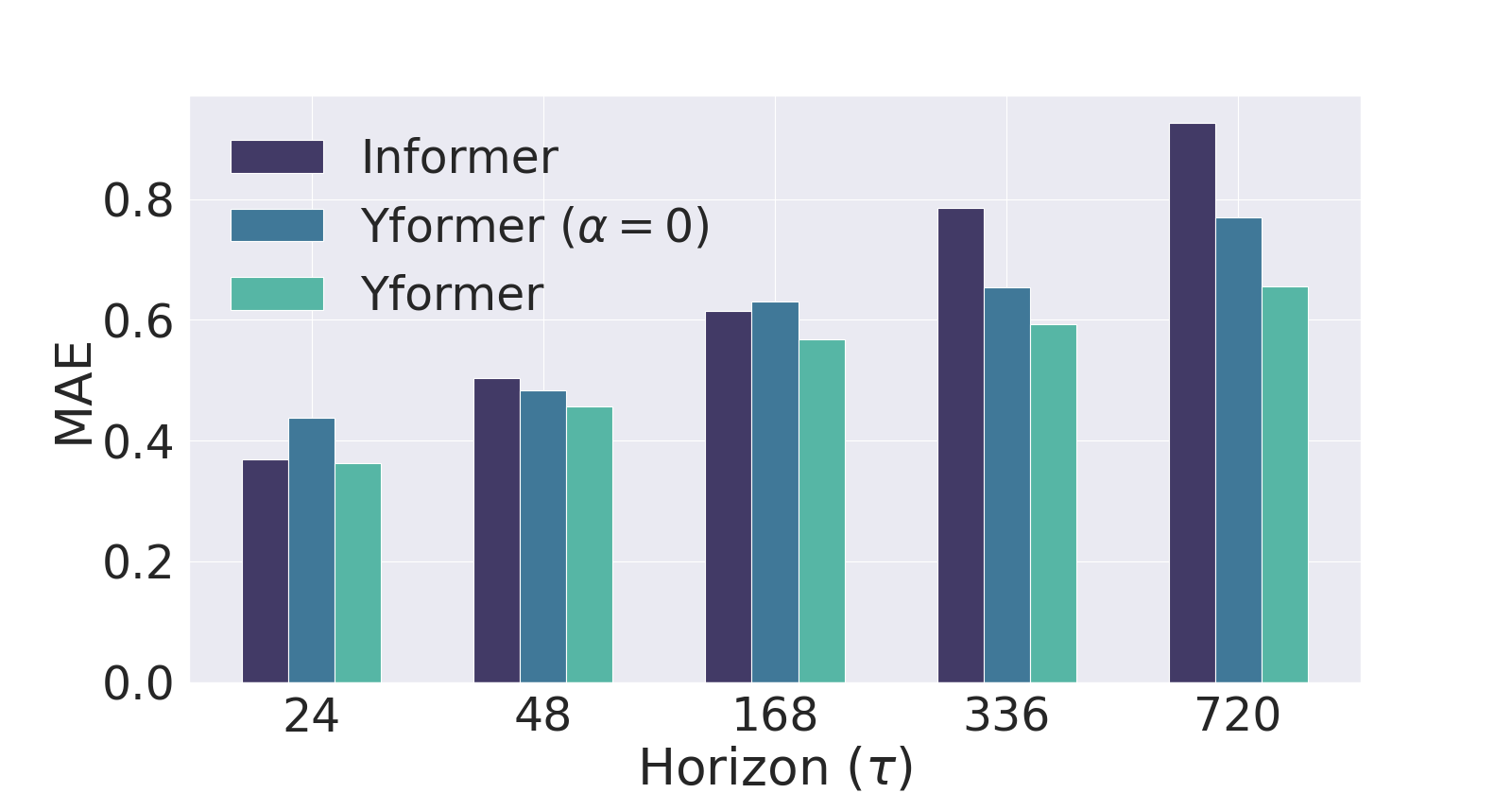}
    }

    \\
    \subfloat[ETTm1 Univariate\label{fig:skipless_ablation_uni}]{%
      \includegraphics[width=0.40\textwidth]{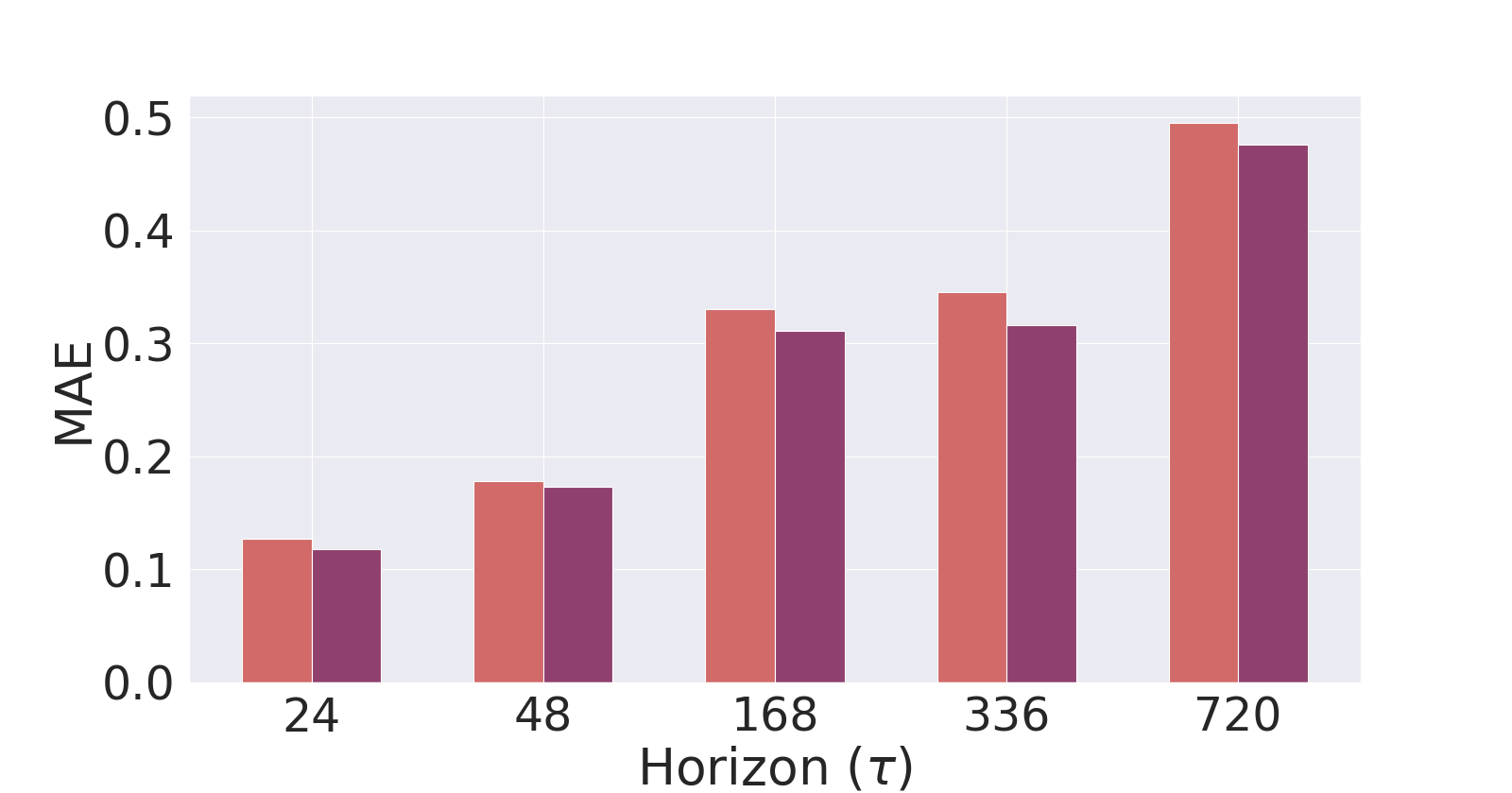}
    }
    \subfloat[ETTm1 Multivariate\label{fig:skipless_ablation_multi}]{%
      \includegraphics[width=0.40\textwidth]{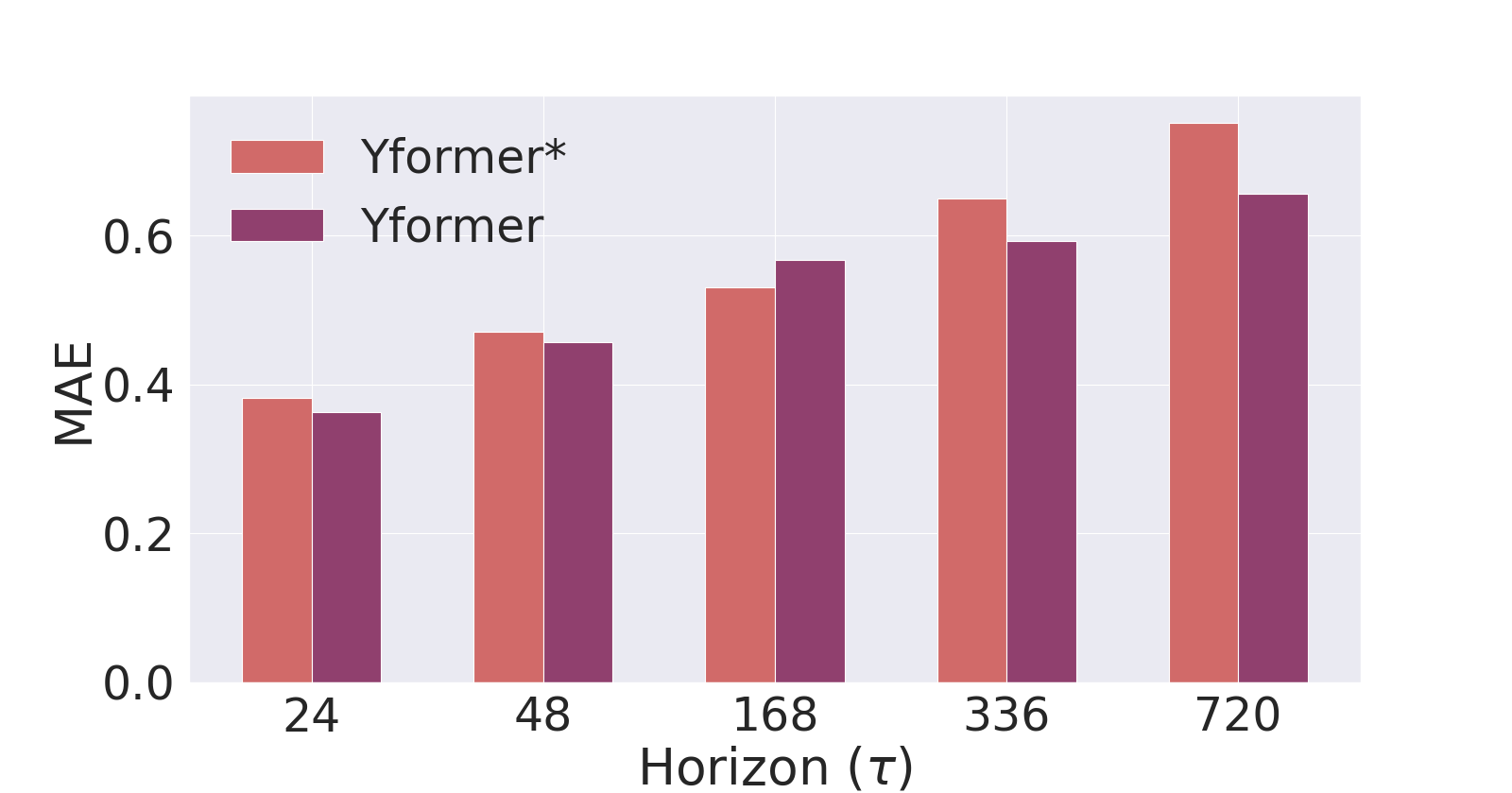}
    }
    \end{tabular}
\caption{(top) Figures \ref{fig:ablation_archi_uni}, \ref{fig:ablation_archi_multi} illustrates the reduction in MAE loss (y-axis) by  the Yformer architecture in comparison with the Informer baseline for the univariate and multivariate settings respectively. The Yformer ($\alpha=0$) represent the Yformer architecture without the reconstruction loss. (bottom) Figures \ref{fig:skipless_ablation_uni}, \ref{fig:skipless_ablation_multi} demonstrate the reduction in MAE loss (y-axis) brought by the addition of U-Net based skip connections (Yformer) to the Yformer architecture without the skip connections (Yformer$^*$). }
\label{fig:ablation_archi}
\end{figure}

In this section, we attempt to understand (1) the improvement brought about by the Y-shaped model architecture, and (2) the impact of the reconstruction loss on the superiority of the Yformer model. Firstly, Figure \ref{fig:model_size_comparison} compares the model complexity for the proposed Yformer model with the Informer baseline model and demonstrates the advantage offered by the Yformer model for longer horizons. Secondly, Figures \ref{fig:ablation_archi_uni}, \ref{fig:ablation_archi_multi}, show that the Yformer architecture performs better or is comparable to the Informer throughout the entire horizon range. Moreover, for the larger horizons, the Yformer architecture without the reconstruction loss i.e. $\alpha=0$, has a clear advantage over the Informer baseline. We attribute this improvement in performance to the additional direct U-Net inspired connections within the Yformer architecture. Using feature maps at multiple resolutions offers a clear advantage by eliminating vanishing gradients and encouraging feature reuse. Figures \ref{fig:ablation_archi_uni}, \ref{fig:ablation_archi_multi} also clearly delineates the advantage offered by adding reconstruction loss as an auxiliary task for the model, by comparing Yformer with Yformer ($\alpha=0$) results. Such a multi-task approach offers regularization to the model by learning parameters that do not overfit on the future target distribution and propels the gradients towards a general distribution that can predict the history along with the future time steps.

\subsection{Effectiveness of the U-Net based skip connections}

To analyze the impact of U-Net based skip-connections, we conduct an ablation study on the Y-former architecture by removing the U-Net skip connections from the encoder to the decoder. We denote this model as Yformer$^*$. Figures \ref{fig:skipless_ablation_uni}, \ref{fig:skipless_ablation_multi} provides a summary of the results obtained after hyperparameter tuning the Yformer$^*$ and comparing it with the proposed Yformer model. The skip connections from the encoder to the decoder improve the performance throughout the entire horizon range for the multivariate setting and offers partial improvement for the univariate setting. Within the multivariate setting, the skip connections have a considerable impact on larger horizons and a smaller impact on the shorter horizons. This observation can be reasoned by considering the fact that long-range forecasting can utilize the additional multi-resolution encoder feature maps encoded by the U-Net based skip connections. Similar reason can be applied to the fact that U-Net based skip connections improve the performance of the multivariate setting more than that of the univariate settings.

\subsection{Reconstruction Factor}
\begin{figure}[t]
    \centering
    \begin{tabular}{c}
    \subfloat[best $\alpha$'s for Univariate\label{fig:alpha_ablation_uni}]{%
      \includegraphics[width=0.33\textwidth]{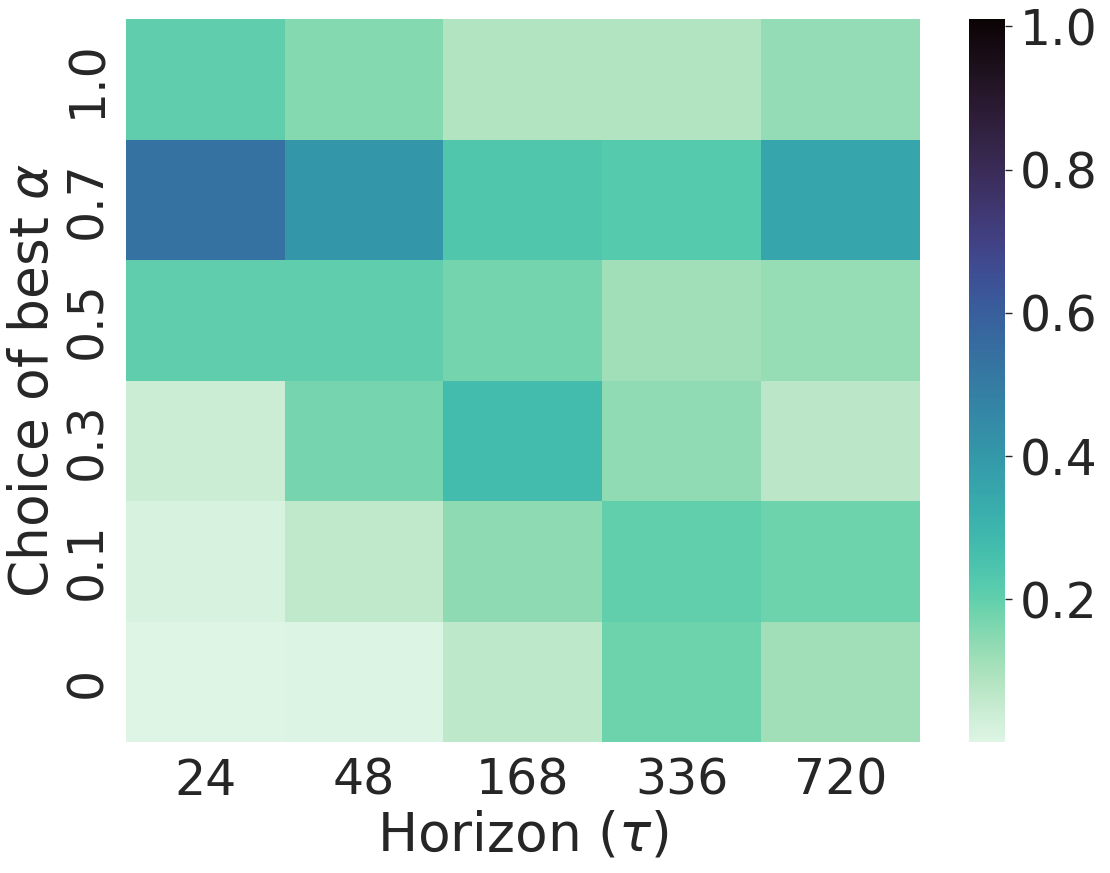}
    }
    \subfloat[best $\alpha$'s for Multivariate\label{fig:alpha_ablation_multi}]{%
      \includegraphics[width=0.33\textwidth]{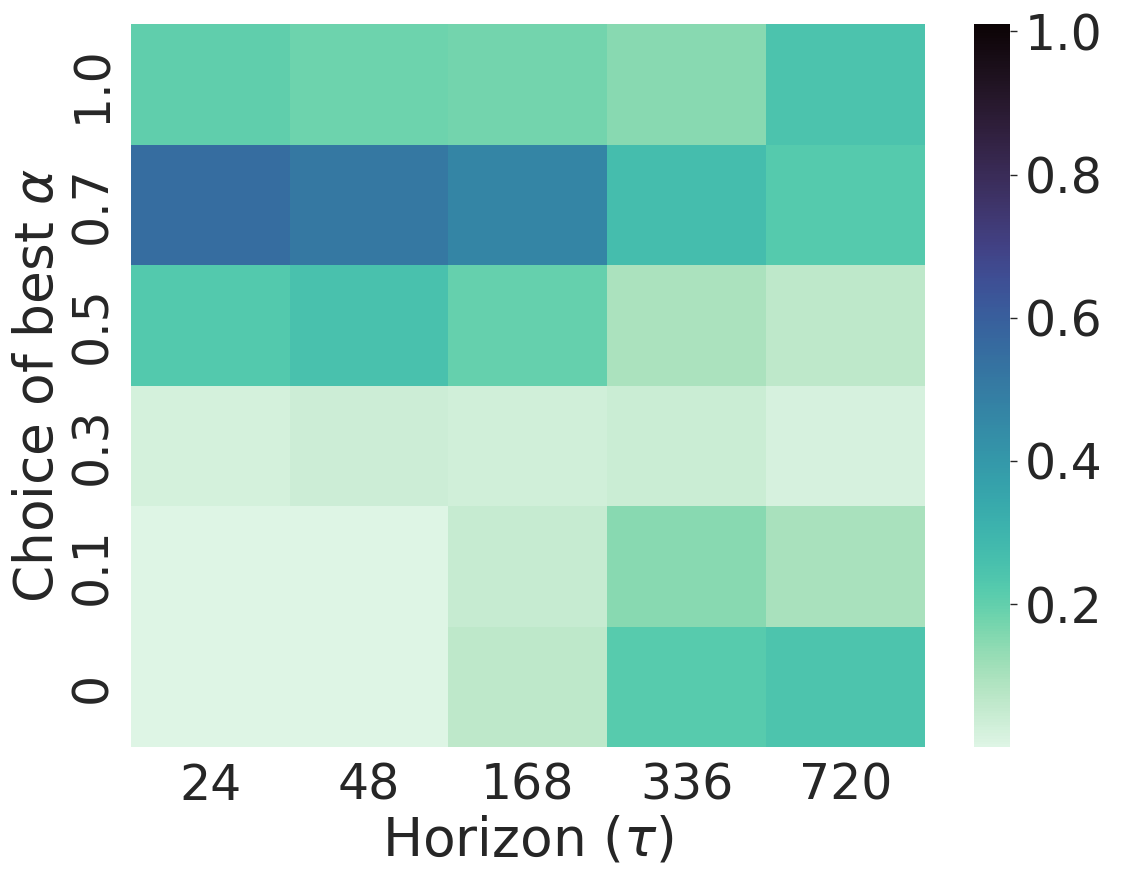}
    }
    \subfloat[Model complexity\label{fig:model_size_comparison}]{%
      \includegraphics[width=0.34\textwidth]{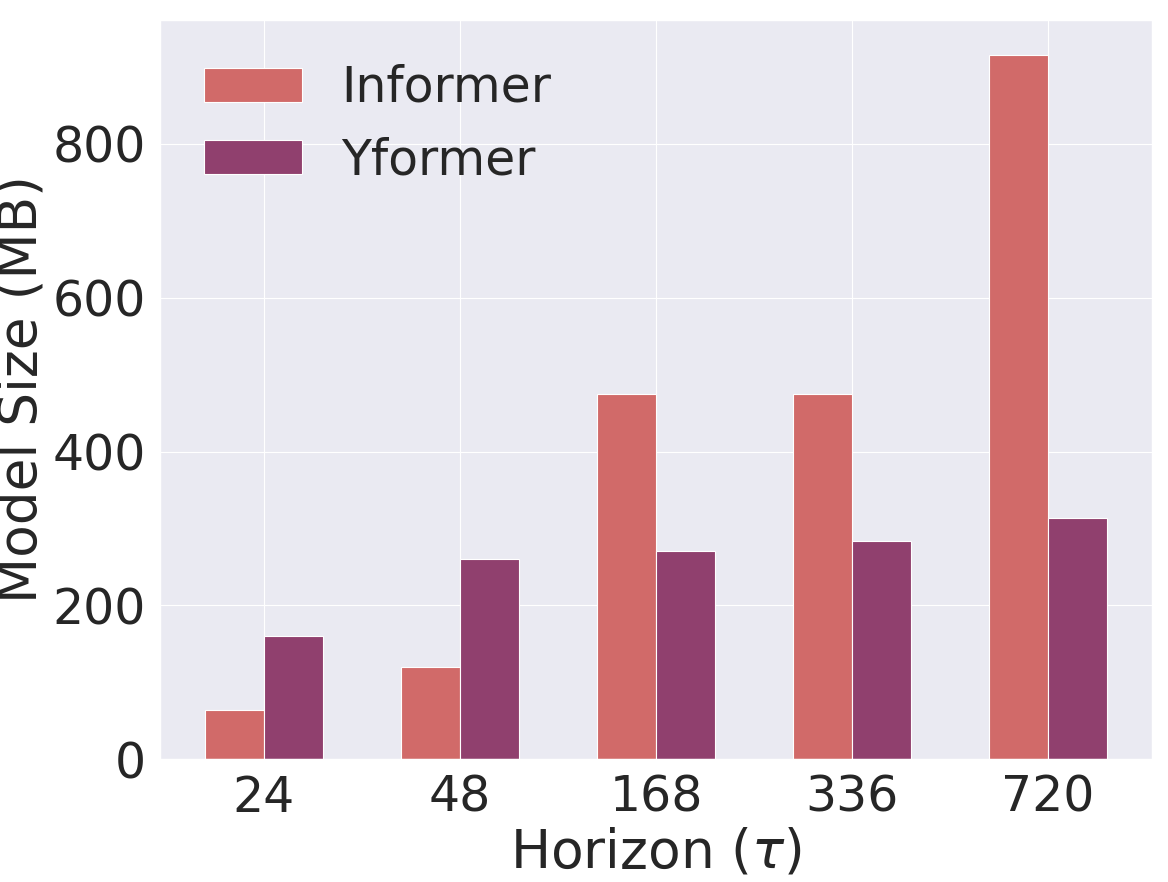}
    }
    \end{tabular}
\caption{Figures \ref{fig:alpha_ablation_uni} and \ref{fig:alpha_ablation_multi} illustrates the distribution of selected Reconstruction factor (y-axis) across the multiple horizons (x-axis). Figure \ref{fig:model_size_comparison}, compares the model size complexity (y-axis) for the multivariate setting across the multiple horizons (x-axis) for the Informer and the Yformer model.}
\label{fig:alpha_ablation}
\end{figure}

How impactful is the reconstruction factor $\alpha$ from the proposed loss in Eq. \ref{eqn:reconstruction}? We aggregated the optimal value chosen by hyperparameter tuning $\alpha$ across different datasets and summarized the distribution in Figures \ref{fig:alpha_ablation_uni} and \ref{fig:alpha_ablation_multi}. 
Interestingly, $\alpha$ value of $0.7$ is the predominant optimal setting across most horizons. Consequently, this shows that a high weight for the reconstruction loss helps the Yformer to achieve a lower loss for the future targets. Moreover, we can observe a trend that $\alpha$ is on average larger for short forecasting horizons signifying the importance of auxiliary loss for the shorter horizons. One possible reason could be that the reconstruction loss generalizes the output distribution better and avoids overfitting on short-horizon lengths. For the longer horizon forecasts, optimal $\alpha$ values are distributed on the lower and upper range of $\alpha$'s evenly, indicating that for long horizons, the reconstruction loss from long history helps for some datasets and does not for other datasets. This could be a characteristic of the dataset having a domain shift within the forecast horizon.

%% file: sections/conclusion.tex
\section{Conclusion}

Time series forecasting is an important business and research problem that has a broad impact in today's world. This paper proposes a novel Y-shaped architecture, specifically designed for the far horizon time series forecasting problem. The study shows the importance of direct connections from the multi-resolution encoder to the decoder and reconstruction loss for the task of time series forecasting. The Yformer couples the U-Net architecture from the image segmentation domain on a sparse transformer model and empirically demonstrates superior performance across multiple datasets for both univariate and multivariate settings. We believe that our work provides a base for future research in the direction of using efficient U-Net based skip connections and the use of reconstruction loss as an auxiliary loss within the time series forecasting community.

\subsubsection{Acknowledgements}: This work was supported by the Federal Ministry for Economic Affairs and Climate Action (BMWK), Germany, within the framework of the IIP-Ecosphere project (project number: 01MK20006D)

%% file: sections/appendix.tex
\section{Appendix : Analysis}

\subsection{Additional ablation results on ETTh2 dataset}
\label{appendix:ablation_etth2}

\begin{figure}[!ht]
    \centering
    \subfloat[ETTh2 Univariate\label{fig:ablation_archi_uni_etth2}]{%
      \includegraphics[width=0.40\textwidth]{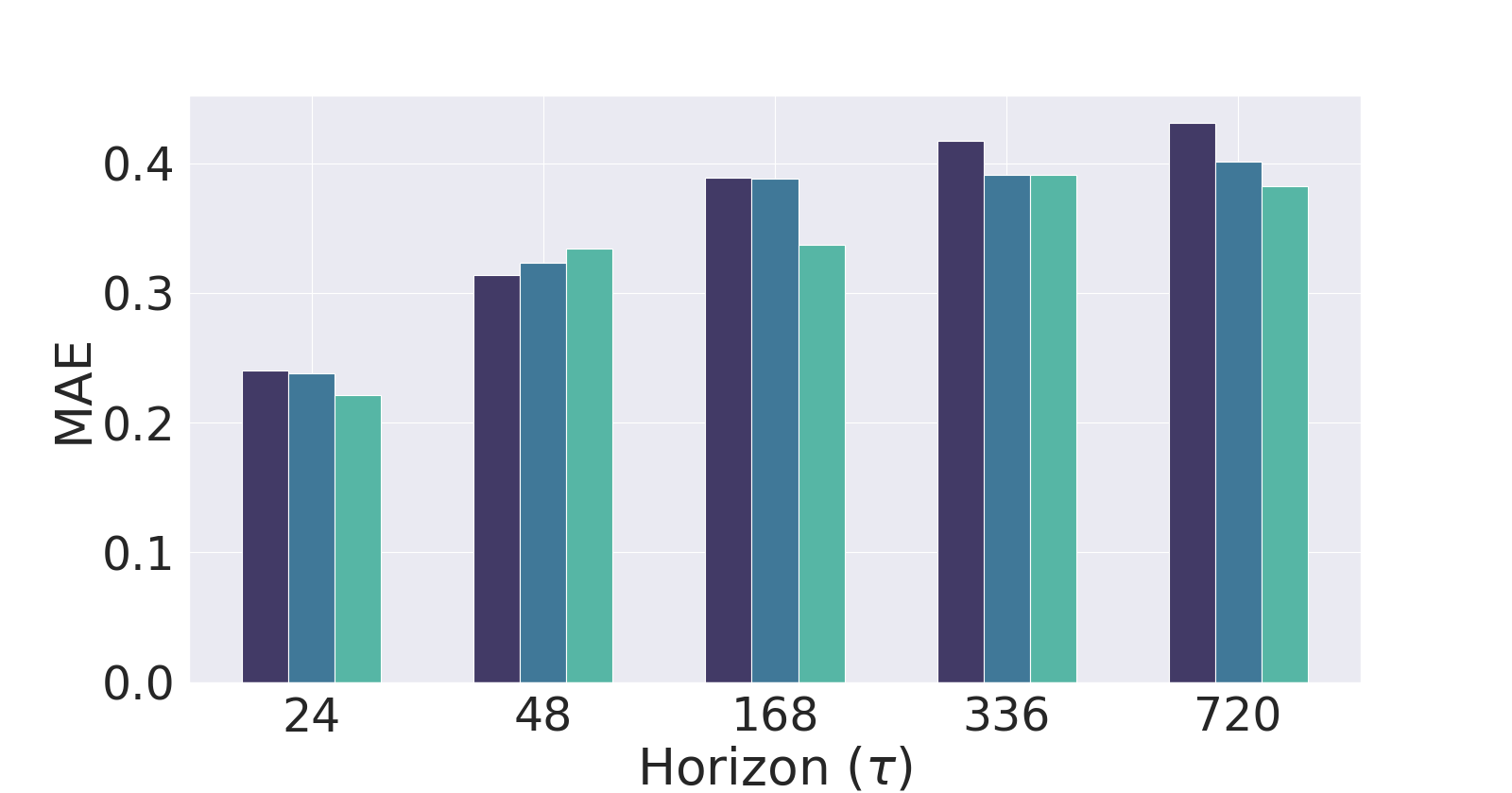}
    }
    \subfloat[ETTh2 Multivariate\label{fig:ablation_archi_multi_etth2}]{%
      \includegraphics[width=0.40\textwidth]{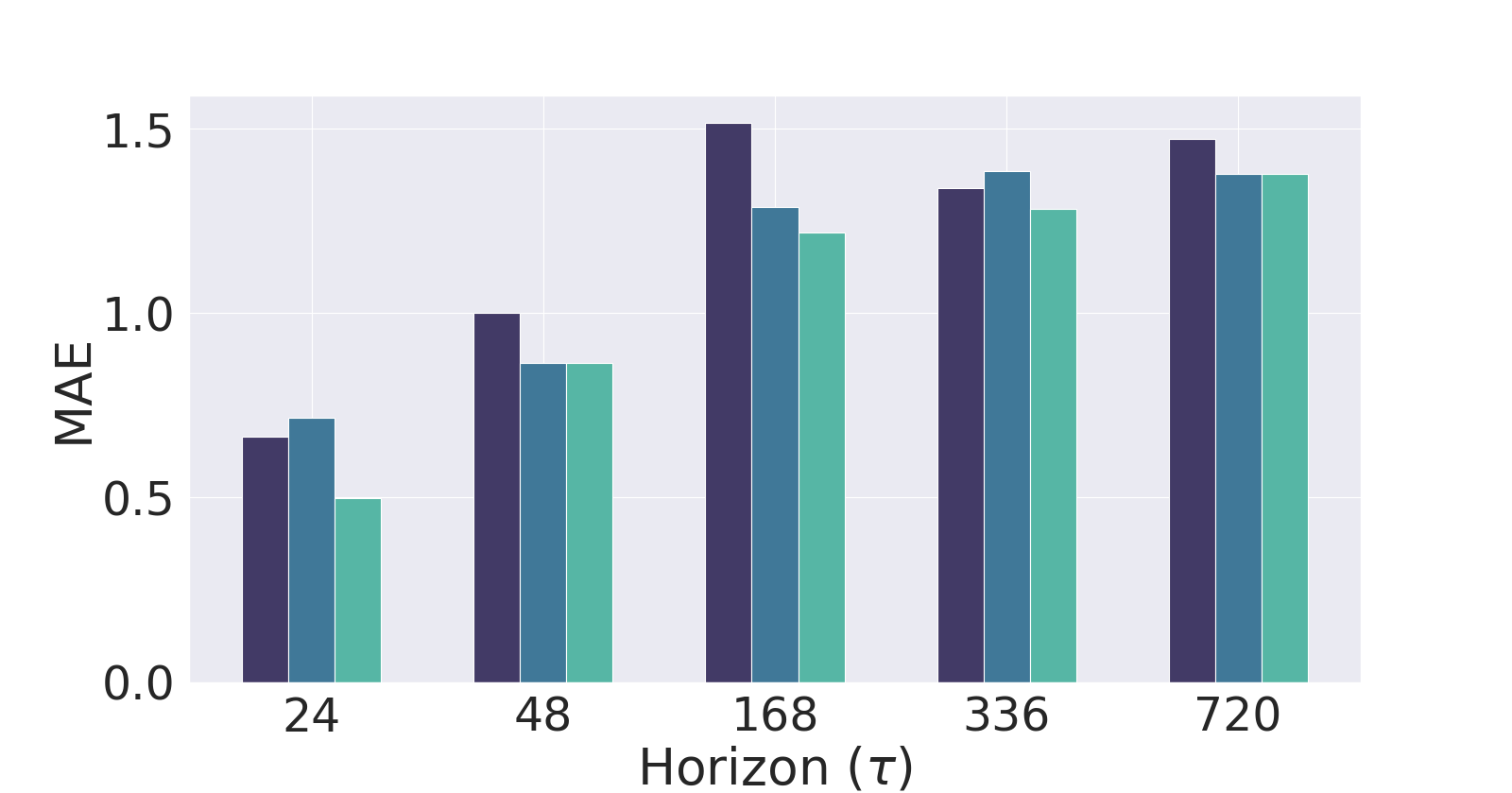}
    } 
\caption{Figures \ref{fig:ablation_archi_multi_etth2}, \ref{fig:ablation_archi_uni_etth2} illustrates the reduction in MAE loss (y-axis) by  the Yformer architecture in comparison with the Informer baseline for the ETTh2 univariate and multivariate settings respectively. The Yformer ($\alpha=0$) represent the Yformer architecture without the reconstruction loss
}
\label{fig:archi_abltation_etth2}
\end{figure}

\begin{figure}[!ht]
    \centering
    \subfloat[ETTh2 Univariate\label{fig:skipless_ablation_uni_ETTh2}]{%
      \includegraphics[width=0.40\textwidth]{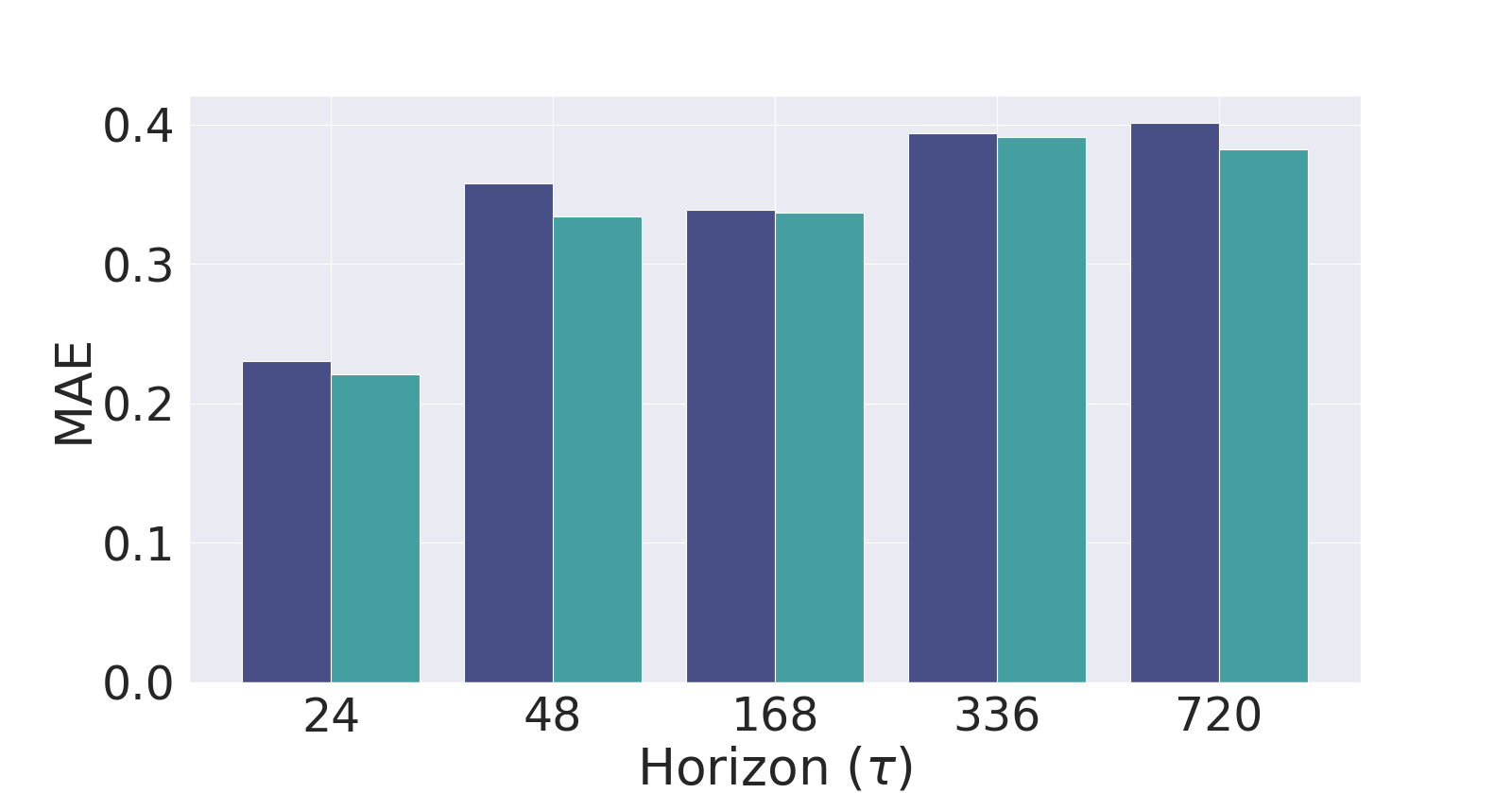}
    }
    \subfloat[ETTh2 Multivariate\label{fig:skipless_ablation_multi_ETTh2}]{%
      \includegraphics[width=0.40\textwidth]{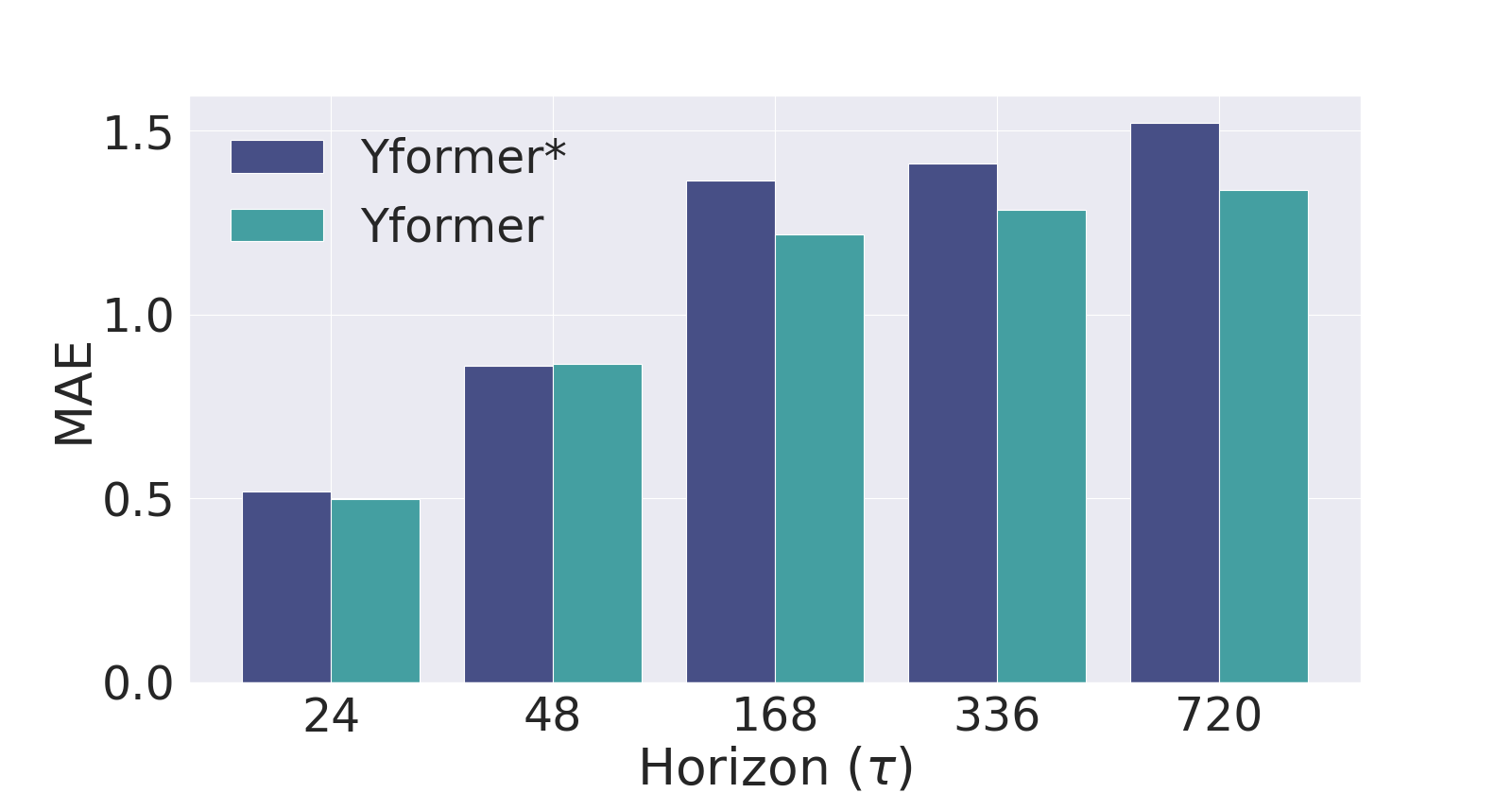}
    } 
\caption{Impact of the U-Net connection for the Yformer architecture. The Yformer$^*$ architecture represents the Yformer without the U-Net connection.}
\label{fig:skipless_ablation_2}
\end{figure}

\subsection{Performance variability analysis}

We report the standard deviation values from the multiple Yformer runs for the ETTh2 dataset and compare them with the numbers reported from the Informer baseline \cite{zhou2020informer}. The standard deviation values are quite small across the three runs of the Yformer with multiple initial seed settings illustrating the stability of Yformer across the multiple horizons.

\begin{table}[htbp!]
\caption{Comparison of Yformer model with the second best performing Informer model for performance variability analysis.}
\resizebox{1\textwidth}{!}{%
\begin{tabular}{|c|c|c|c|c|c|c|c|}
\hline
\multicolumn{1}{|c|}{Setting} & \multicolumn{1}{c|}{Model} & Metric & \multicolumn{1}{c|}{24} & \multicolumn{1}{c|}{48} & \multicolumn{1}{c|}{168} & \multicolumn{1}{c|}{336} & \multicolumn{1}{c|}{720} \\ \hline
\multirow{4}{*}{Univariate} & \multirow{2}{*}{Yformer}  & MSE    & $0.082\pm0.004$ & $0.172\pm0.016$ & $0.174\pm0.009$ & $0.224\pm0.038$ & $0.211\pm0.005$ \\ \cline{3-8} 
                                    &                           & MAE    & $0.221\pm0.006$ & $0.334\pm0.014$ & $0.337\pm0.007$ & $0.391\pm0.036$ & $0.382\pm0.005$ \\ \cline{2-8} 
                                    & \multirow{2}{*}{Informer} & MSE    & 0.093         & 0.155         & 0.232         & 0.263         & 0.277         \\ \cline{3-8} 
                                    &                           & MAE    & 0.24          & 0.314         & 0.389         & 0.417         & 0.431         \\ \hline
\multirow{4}{*}{Multivariate}        & \multirow{2}{*}{Yformer}   & MSE    & $0.412\pm0.063$             & $1.171\pm0.027$           & $2.171\pm0.105$            & $2.260\pm0.112$            & $2.595\pm0.131$              \\ \cline{3-8} 
                              &                            & MAE    & $0.498\pm0.049$             & $0.865\pm0.029$           & $1.218\pm0.047$            & $1.283\pm0.009$            & $1.337\pm0.066$              \\ \cline{2-8} 
                              & \multirow{2}{*}{Informer}  & MSE    & 0.720                    & 1.457                   & 3.489                    & 2.723                    & 3.467                    \\ \cline{3-8} 
                              &                            & MAE    & 0.665                   & 1.001                   & 1.515                    & 1.340                     & 1.473                    \\ \hline
\end{tabular}%
}
\end{table}

\newpage

\section{Appendix: Operators}

$\operatorname{\textbf{ProbSparseAttn}}$: Attention module that uses the ProbSparse method introduced in \cite{zhou2020informer}. The query matrix $\overline{\boldsymbol{Q}} \in \mathbb{R}^{L_Q \times d}$ denotes the sparse query matrix with $u$ dominant queries.

\begin{equation}
\begin{aligned}
  \mathcal{A^{\text{PropSparse}}}(\boldsymbol{\overline{Q}}, \boldsymbol{K}, \boldsymbol{V}) &= \text{Softmax}(\frac{\boldsymbol{\overline{Q}}\boldsymbol{K}^T}{\sqrt{d}})\boldsymbol{V}
\end{aligned}
\label{eqn:probattn}
\end{equation}

$\operatorname{\textbf{MaskedAttn}}$: Canonical self-attention with masking to prevent positions from attending to subsequent positions in the future \cite{vaswani2017attention}.

$\operatorname{\textbf{Conv1d}}$: Given $N$ batches of 1D array of length $L$ and $C$ number of channels/dimensions. A convolution operation produces an output: 

\begin{equation}
\begin{aligned}
    \text{out}(N_i, C_{\text{out}_j}) = \text{bias}(C_{\text{out}_j}) +
        \sum_{k = 0}^{C_{in} - 1} \text{weight}(C_{\text{out}_j}, k)
        \star \text{input}(N_i, k)
\end{aligned}
\label{eqn:conv1d}
\end{equation}

For further reference please visit \href{https://pytorch.org/docs/stable/generated/torch.nn.Conv1d.html}{pytorch Conv1D} page

$\operatorname{\textbf{LayerNorm}}$: Layer Normalization introduced in \cite{layernorm}, normalizes the inputs across channels/dimensions. $\operatorname{LayerNorm}$ is the default normalization in common transformer architectures \cite{vaswani2017attention}. Here, $\gamma$ and $\beta$ are learnable affine transformations.

\begin{equation}
\begin{aligned}
    \text{out}(N, *) = \frac{\text{input}(N, *) - \mathrm{E}[\text{input}(N, *)]}{ \sqrt{\mathrm{Var}[\text{input}(N, *) ] + \epsilon}} * \gamma + \beta
\end{aligned}
\label{eqn:layernorm}
\end{equation}

$\operatorname{\textbf{MaxPool}}$: Given $N$ batches of 1D array of length $L$, and $C$ number of channels/dimensions. A $\operatorname{MaxPool}$ operation produces an output. 

\begin{equation}
\begin{aligned}
    \text{out}(N_i, C_j, k) = \max_{m=0, \ldots, \text{kernel\_size} - 1}
                \text{input}(N_i, C_j, \text{stride} \times k + m)
\end{aligned}
\label{eqn:maxpool}
\end{equation}

For further reference please visit \href{https://pytorch.org/docs/stable/generated/torch.nn.MaxPool1d.html}{pytorch MaxPool1D} page 

$\operatorname{\textbf{ELU}}$: Given an input $x$, the $\operatorname{ELU}$ applies element-wise non linear activation function as shown.

\begin{equation}
\begin{aligned}
    \text{ELU}(x) = \begin{cases}
        x, & \text{ if } x > 0\\
        \alpha * (\exp(x) - 1), & \text{ if } x \leq 0
        \end{cases}
\end{aligned}
\label{eqn:elu}
\end{equation}

$\operatorname{\textbf{ConvTranspose1d}}$: Also known as deconvolution or fractionally strided convolution, uses convolution on padded input to produce upsampled outputs (see \href{https://pytorch.org/docs/stable/generated/torch.nn.ConvTranspose1d.html}{pytorch ConvTranspose1d} page).

\section{Appendix : Hyperparameters}
\label{appendix:hyperparameters}

We follow Informer \cite{zhou2020informer} baseline for all the hyperparameter setting like the convolution kernel size, stride etc. The hyperparameter tuning performed are only for the parameters mentioned below. In order to reproduce the experiments, please use the default Informer/Yformer configurations and adapt only the below mentioned parameters for each horizon.

\begin{table}[htbp!]
\caption{Optimal hyperparameters across different horizon and datasets for the univariate setting. All the remaining hyperparameters are retained from the Informer Model.}
\label{tbl:hyp_univariate}
\centering
\resizebox{1\textwidth}{!}{%
\begin{tabular}{|c|c|c|S|S|S|c|c|}
\hline
Dataset                & Horizon $\tau$& History Length & {Weight Decay} & {Learning Rate} & {Reconstruction Factor $\alpha$} & Batch Size & Encoder Blocks \\ \hline
\multirow{5}{*}{ETTh1} & 24      & 720        & 0            & 0.0001        & 0.7   & 32         & 2              \\ \cline{2-8} 
                      & 48      & 720        & 0         & 0.0001         & 0.7   & 16         & 4              \\ \cline{2-8} 
                      & 168     & 720        & 0            & 0.001         & 0.7   & 32         & 4              \\ \cline{2-8} 
                      & 336     & 720        & 0.05         & 0.0001        & 0.1   & 32         & 4              \\ \cline{2-8} 
                      & 720     & 720        & 0.05         & 0.0001        & 0.7   & 16         & 2              \\ \hline
\multirow{5}{*}{ETTh2} & 24      & 48         & 0            & 0.0001        & 0.7   & 32         & 2              \\ \cline{2-8} 
                      & 48      & 96         & 0.02         & 0.0001        & 0.3   & 32         & 4              \\ \cline{2-8} 
                      & 168     & 336        & 0.02         & 0.001         & 0.3   & 32         & 2              \\ \cline{2-8} 
                      & 336     & 336        & 0.09         & 0.0001        & 0     & 32         & 2              \\ \cline{2-8} 
                      & 720     & 336        & 0.09         & 0.0001        & 0.7   & 16         & 2              \\ \hline
\multirow{5}{*}{ETTm1} & 24      & 96         & 0.02         & 0.0001        & 0.7   & 32         & 4              \\ \cline{2-8} 
                      & 48      & 96         & 0.02         & 0.0001        & 0.7   & 32         & 4              \\ \cline{2-8} 
                      & 96      & 384        & 0.02         & 0.0001        & 0.1   & 32         & 4              \\ \cline{2-8} 
                      & 288     & 384        & 0.02         & 0.001         & 0.7   & 16         & 2              \\ \cline{2-8} 
                      & 672     & 384        & 0.07         & 0.001         & 0.3   & 16         & 2              \\ \hline
\multirow{5}{*}{ECL}   & 48      & 168        & 0            & 0.0001        & 0.7   & 16         & 2              \\ \cline{2-8} 
                      & 168     & 168        & 0.01         & 0.0001        & 0.3   & 16         & 2              \\ \cline{2-8} 
                      & 336     & 168        & 0.01         & 0.0001        & 0.7   & 16         & 2              \\ \cline{2-8} 
                      & 720     & 168        & 0            & 0.0001        & 0.1   & 16         & 2              \\ \cline{2-8} 
                      & 960     & 48         & 0            & 0.0001        & 0.5   & 16         & 4              \\ \hline
\end{tabular}%
}

\end{table}

\begin{table}[htbp!]
\caption{Optimal hyperparameters across different horizon and datasets for the multivariate setting. All the remaining hyperparameters are retained from the Informer Model.}
\label{tbl:hyp_multivariate}
\centering
\resizebox{1\textwidth}{!}{%
\begin{tabular}{|c|c|c|S|S|S|c|c|}
\hline
Dataset                & Horizon $\tau$& History Length & {Weight Decay} & {Learning Rate} & {Reconstruction Factor $\alpha$} & Batch Size & Encoder Blocks \\ \hline
\multirow{5}{*}{ETTh1} & 24      & 48         & 0            & 0.0001        & 0.7   & 32         & 3              \\ \cline{2-8} 
                      & 48      & 96         & 0.02         & 0.001         & 0.5   & 32         & 2              \\ \cline{2-8} 
                      & 168     & 168        & 0.02         & 0.001         & 0.7   & 32         & 2              \\ \cline{2-8} 
                      & 336     & 168        & 0            & 0.0001        & 0.7   & 32         & 4              \\ \cline{2-8} 
                      & 720     & 336        & 0.05         & 0.0001        & 1     & 16         & 2              \\ \hline
\multirow{5}{*}{ETTh2} & 24      & 48         & 0            & 0.0001        & 0.7   & 32         & 2              \\ \cline{2-8} 
                      & 48      & 96         & 0.02         & 0.001         & 0     & 32         & 4              \\ \cline{2-8} 
                      & 168     & 336        & 0.09         & 0.001         & 0.7   & 32         & 2              \\ \cline{2-8} 
                      & 336     & 336        & 0.07         & 0.001         & 0.3   & 32         & 2              \\ \cline{2-8} 
                      & 720     & 336        & 0            & 0.0001        & 0     & 16         & 2              \\ \hline
\multirow{5}{*}{ETTm1} & 24      & 672        & 0            & 0.0001        & 0.7   & 32         & 2              \\ \cline{2-8} 
                      & 48      & 96         & 0            & 0.0001        & 0.7   & 32         & 4              \\ \cline{2-8} 
                      & 96      & 384        & 0.05         & 0.0001        & 0.7   & 32         & 4              \\ \cline{2-8} 
                      & 288     & 672        & 0.02         & 0.001         & 0.5   & 16         & 2              \\ \cline{2-8} 
                      & 672     & 672        & 0.02         & 0.0001        & 0.3   & 16         & 2              \\ \hline
\multirow{5}{*}{ECL}   & 48      & 24         & 0            & 0.0001        & 0.7   & 16         & 3              \\ \cline{2-8} 
                      & 168     & 48         & 0            & 0.0001        & 0.7   & 16         & 3              \\ \cline{2-8} 
                      & 336     & 24         & 0            & 0.0001        & 0.5   & 16         & 2              \\ \cline{2-8} 
                      & 720     & 48         & 0            & 0.0001        & 0.7   & 16         & 2              \\ \cline{2-8} 
                      & 960     & 336        & 0            & 0.0001        & 0.7   & 16         & 2              \\ \hline
\end{tabular}%
}

\end{table}